\documentclass[sigconf]{acmart}

\AtBeginDocument{%
  \providecommand\BibTeX{{%
    \normalfont B\kern-0.5em{\scshape i\kern-0.25em b}\kern-0.8em\TeX}}}

\setcopyright{acmlicensed}
\copyrightyear{2024}
\acmYear{2024}
\setcopyright{rightsretained}
\acmConference[KDD '24]{Proceedings of the 30th ACM SIGKDD Conference on Knowledge Discovery and Data Mining}{August 25--29, 2024}{Barcelona, Spain}
\acmBooktitle{Proceedings of the 30th ACM SIGKDD Conference on Knowledge Discovery and Data Mining (KDD '24), August 25--29, 2024, Barcelona, Spain}\acmDOI{10.1145/3637528.3671590}
\acmISBN{979-8-4007-0490-1/24/08}

\makeatletter
\gdef\@copyrightpermission{
 \begin{minipage}{0.3\columnwidth}
 \href{https://creativecommons.org/licenses/by/4.0/}{\includegraphics[width=0.90\textwidth]{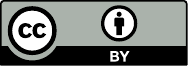}}
 \end{minipage}\hfill
 \begin{minipage}{0.7\columnwidth}
 \href{https://creativecommons.org/licenses/by/4.0/}{This work is licensed under a Creative Commons
Attribution International 4.0 License.}
 \end{minipage}
 \vspace{5pt}
}
\makeatother

\usepackage{amsmath,amsfonts,bm}

\def\eqref#1{equation~\ref{#1}}

\def\1{\bm{1}}

\DeclareMathAlphabet{\mathsfit}{\encodingdefault}{\sfdefault}{m}{sl}
\SetMathAlphabet{\mathsfit}{bold}{\encodingdefault}{\sfdefault}{bx}{n}

\usepackage{pdfpages}
\usepackage{paralist}  
\usepackage{subcaption}
\usepackage{dblfloatfix}
\usepackage{balance}
\usepackage{algorithm}
\usepackage{algpseudocode}%
\usepackage{multirow}
\usepackage[flushleft]{threeparttable}
\usepackage{booktabs} 
\algrenewcommand\algorithmicrequire{\textbf{Input:}}
\algrenewcommand\algorithmicensure{\textbf{Output:}}
\begin{document}

\title{Distributed Harmonization: Federated Clustered Batch Effect Adjustment and Generalization}

\author{Bao Hoang}
\email{hoangbao@msu.edu}
\affiliation{%
  \institution{Michigan State University}
  \city{East Lansing}
  \state{Michigan}
  \country{USA}
  \postcode{48824}
}

\orcid{0009-0002-3462-7656}

\author{Yijiang Pang}
\email{pangyiji@msu.edu}
\affiliation{%
  \institution{Michigan State University}
  \city{East Lansing}
  \state{Michigan}
  \country{USA}
  \postcode{48824}
}

\author{Siqi Liang}
\email{liangsi4@msu.edu}
\affiliation{%
  \institution{Michigan State University}
  \city{East Lansing}
  \state{Michigan}
  \country{USA}
  \postcode{48824}
}

\author{Liang Zhan}
\email{liang.zhan@pitt.edu}
\affiliation{%
  \institution{University of Pittsburgh}
  \city{Pittsburgh}
  \state{Pennsylvania}
  \country{USA}
  \postcode{15260}
}

\author{Paul M. Thompson}
\email{pthomp@usc.edu}
\affiliation{%
  \institution{University of Southern California}
  \city{Los Angeles}
  \state{California}
  \country{USA} 
  \postcode{90007}
}

\author{Jiayu Zhou}
\authornote{Corresponding author}
\email{jiayuz@msu.edu}
\affiliation{%
  \institution{Michigan State University}
  \city{East Lansing}
  \state{Michigan}
  \country{USA}
  \postcode{48824}
}

\renewcommand{\shortauthors}{Bao Hoang et al.}

\begin{abstract}
Independent and identically distributed (\emph{i.i.d.}) data is essential to many data analysis and modeling techniques. In the medical domain, collecting data from multiple sites or institutions is a common strategy that guarantees sufficient clinical diversity, determined by the decentralized nature of medical data. However, data from various sites are easily biased by the local environment or facilities, thereby violating the \emph{i.i.d.} rule. A common strategy is to harmonize the site bias while retaining important biological information. The \textsc{ComBat} is among the most popular harmonization approaches and has recently been extended to handle distributed sites. However, when faced with situations involving newly joined sites in training or evaluating data from unknown/unseen sites, \textsc{ComBat} lacks compatibility and requires retraining with data from all the sites. The retraining leads to significant computational and logistic overhead that is usually prohibitive. In this work, we develop a novel \textit{Cluster ComBat} harmonization algorithm, which leverages cluster patterns of the data in different sites and greatly advances the usability of \textsc{ComBat} harmonization. We use extensive simulation and real medical imaging data from ADNI to demonstrate the superiority of the proposed approach. Our codes are provided in \href{https://github.com/illidanlab/distributed-cluster-harmonization}{https://github.com/illidanlab/distributed-cluster-harmonization}.
\end{abstract}

\ccsdesc[500]{Applied computing~Imaging}
\ccsdesc[500]{Computing methodologies~Distributed algorithms}
\ccsdesc[500]{Computing methodologies~Machine learning}
\keywords{Harmonization, Distributed Algorithm, Neuroimaging, Medical Data}



\maketitle
\section{Introduction}

The recent advances in machine learning approaches have greatly advanced biomedical data analysis. In brain imaging analysis, for example, Magnetic Resonance Imaging (MRI) has been used for the detection and disease progression of many diseases, such as Mild Cognitive Impairment (MCI)~\cite{Wang2018,zhou2011multi,zhou2013modeling, zhou2012modeling}, Parkinson's disease~\cite{GarciaSantaCruz2023}, and Brain Tumor Detection~\cite{Abdusalomov2023}. However, one critical challenge with brain imaging is that the brain imaging is sensitive to scanner or protocol effect~\cite{zhan2014understanding,Pfaehler2019},  also commonly referred to as site effect or batch effect, leading to the fact that brain imaging from multiple sites is not independent and identically distributed (\emph{i.i.d.}). The bias in non-\emph{i.i.d.} data will cause unstable prediction performance and poor generalization performance to unseen data~\cite{Yang}. Consequently, developing an algorithm that can eliminate these types of bias ensures consistent and reliable outcomes in the deployment of machine learning models within the medical imaging domain. 

\textsc{ComBat}~\cite{Johnson2006} is a well-known harmonization technique and has been shown to be helpful in mitigating the site effect of neuroimaging data introduced by multiple sites sampling~\cite{Orlhac2021}. Despite its utility, one of the central ideas is that \textsc{ComBat} debias the site effect independently according to the (local) site data, which induces one critical limitation, its inability to evaluate site effects coming from unseen or unknown sites without undergoing a retraining process. The requirement of retraining is hindered by substantial computational costs when it comes to real-world deployment, especially when dealing with large datasets from multiple sites. This limitation underscores the need for a more efficient and broadly applicable approach in mitigating the site effects of medical data, especially neuroimaging data.

Furthermore, a centralized setting for site effect harmonization introduces extra concerns. For instance, sharing data directly among multiple sites to apply \textsc{ComBat} harmonization poses challenges to the security of confidential data and the protection of patient privacy. Direct training on all the data is often impractical in the medical domain. This underscores the need to develop harmonization algorithms in a decentralized manner that can effectively harmonize without gathering data from all sites while maintaining competitive performance in the centralized setting. Particularly, the Distributed ComBat~\cite{Chen2022}, a distributed version of the \textsc{ComBat} harmonization algorithm, has demonstrated its harmonization capability meanwhile obeying decentralized manners. Nevertheless, Distributed ComBat suffers from the same limitation as the original \textsc{ComBat}, i.e., it cannot estimate site effects from \emph{unseen sites} without retraining.

Because a significant part of site effects in medical data are ultimately rooted in medical instruments, e.g., MRI scanners from different manufacturers and configurations, the bias underneath the sites may not be independent and may exhibit clustering structures. In this paper, we proposed the \textit{Cluster ComBat} method, an extension of the original \textsc{ComBat} algorithms that leverages the cluster patterns of site effects. This approach enables the estimation of site effects from unknown sites without necessitating a retraining process. Furthermore, we also developed a distributed version \textit{Cluster ComBat} and demonstrated its efficacy in harmonizing data obeying decentralized manners. Our empirical findings show that \textit{Cluster ComBat} in both centralized and decentralized settings outperform their respective counterparts on both synthetic and real-world neuroimaging datasets.

\section{Related Works}

\textbf{Brain Imaging}. 
The integration of brain imaging and machine learning has drawn significant attention in recent research, with a focus on advancing diagnostic capabilities and understanding complex neurological conditions ~\cite{Singh2022, Monsour2022}. Recent research advancements highlight the potential of machine learning techniques in exploring underlying complex patterns within neuroimaging data. For example, T1-weighted MRI with Lasso Regression, a statistical technique \cite{Tibshirani1996}, has proven effective in detecting MCI \cite{Wang2018}. Their findings show promising results in the early detection of MCI, emphasizing the significance of early intervention and treatment. Furthermore, complicated deep learning architectures, such as YOLOv7 \cite{wang2023yolov7}, have also demonstrated exceptional predictive performance in brain tumor detection using T1-weighted MRI \cite{Abdusalomov2023}. Besides, Diffusion Tensor Imaging (DTI) also shows valuable information related to Alzheimer’s disease (AD) pathology and achieves promising performance in the diagnosis and progression modeling of AD using machine learning classification \cite{Zhan2015, Zhan2015-2, Wang2016, Wang2017}. 

\noindent\textbf{Distributed Learning.}  
Preserving the privacy of users' information is a crucial issue that needs to be considered as an important aspect to evaluate in a learning algorithm ~\cite{Li2021, Fukasawa2004, xie2018differentially, hong2022dynamic,hong2021learning}. 
Especially in a healthcare setting, where the health data is sensitive, we need to design a distributed learning approach that avoids leaking any private information from hospitals' data ~\cite{xie2017privacy, Welten2022, Wirth2021, Haque2023}. 
For example, for brain imaging, federated learning, a distributed machine learning algorithm, has proven to be effective in analyzing neuroimaging for cognitive detection tasks while still protecting patients' information ~\cite{Sandhu, dipro2022federated}. 
Moreover, for health records data, to avoid sharing raw data between sites, the use of first-order and second-order gradients of the likelihood function of sites has been shown to be sufficient for achieving high accuracy in classification tasks~\cite{Duan2019}. 
In addition, the distributed version of generalized linear mixed models (GLMM) can also achieve nearly identical results in analyzing electronic health records data as in a centralized setting ~\cite{Yan2023}. 
Relating to harmonization methods in a decentralized setting, Distributed ComBat~\cite{Chen2022} and Federated Learning ComBat~\cite{Silva2023} have been developed to harmonize neuroimaging without the need for sharing information between hospitals.

\noindent\textbf{ComBat Harmonization.} \textsc{ComBat} harmonization, initially designed for applications in bioinformatics and genomics, is an effective strategy in mitigating batch effects or site effects within high-dimensional data \cite{Johnson2006}. 
It has been adopted to address various situations and problems \cite{Bayer2022}. 
Fully Bayesian ComBat~\cite{Reynolds2023} investigates the advantage of using Monte Carlo sampling for statistical inference in harmonization algorithms. 
ComBat-GAM~\cite{Pomponio2020} extends the model's capability by also estimating non-linear effects that came from biological covariates, in contrast to only linear effects considered in the original \textsc{ComBat} model.
Longitudinal ComBat~\cite{Beer2020} is designed for datasets collected over multiple time points from the same subjects, effectively taking into account variations within each subject over time and considering changes in linear covariates. 
To preserve the privacy of brain imaging across multiple hospitals or sites, Distributed ComBat~\cite{Chen2022} introduces a decentralized learning version of the original ComBat, which can also harmonize data in decentralized settings. 
Combining the strengths of ComBat-GAM and Distributed ComBat, Federated Learning ComBat~\cite{Silva2023} not only estimates non-linear effects from biological covariates but also utilizes the FedAvg algorithm~\cite{FedAVG} to protect the privacy of data. However, they are not applicable in large-scale studies when new sites join the analysis after the harmonization (e.g., \cite{schijven2023large,thompson2014enigma}). 

\section{Methods}

\subsection{Preliminary: \textsc{ComBat} Harmonization}

\textsc{ComBat}~\cite{Johnson2006} adjusts the location (mean) and scale (variance) of data from different sites for the requirement of downstream analysis tasks. 
Assume given a dataset with $G$ features collected from $M$ different sites. 
For each site $i \in [M]$, there are $N_i$ samples, and $N=\sum_{i \in [M]}{N_i}$ is the total number of samples.
\textsc{ComBat} follows an L/S model assuming that, for each sample $j \in [N_i]$ on site $i$, the value of $g \in [G]$ feature $y_{ijg}$ can be modeled as:
\begin{align}
    y_{ijg} = \alpha_g + X_{ij}\beta_g + \gamma_{ig} + \delta_{ig}\epsilon_{ijg},
\end{align}
where $\alpha_g$ is the mean value of that feature, $X_{ij}$ is the biological covariates (e.g., age, sex), and $\beta_g$ is the regression coefficient of $X_{ij}$. 
$\gamma_g$ represents additive effects from site $i$, while $\delta_g$ represents the corresponding multiplicative effects. 
Also, the error term $\epsilon_{ijg}$ is assumed to be drawn from a Normal distribution $\mathcal{N}(0, \sigma_g^2)$. 
We call these site-wise effect parameters as \textit{harmonization parameters}.

The L/S model assumes that different sites would have different \emph{site effects} on their own data. 
Thus, removing both additive and multiplicative effects from data within each site is mandatory for the later regression task. 
The empirical Bayes algorithm is typically used to estimate these harmonization parameters in \textsc{ComBat}-related approaches~\cite{Johnson2006,Pomponio2020}.

First, \textsc{ComBat} standardizes the data feature-wise:
\begin{align}
\label{eq:standarize}
    Z_{ijg} = \frac{y_{ijg} - \hat{\alpha}_g - X_{ij}\hat{\beta}_g}{\hat{\sigma}_g},
\end{align}
where $\hat{\sigma}_g^2 = \frac{1}{N}\sum_{ij}(y_{ijg} - \hat{\alpha}_g - X_{ij}\hat{\beta}_g - \hat{\gamma}_{ig})^2$, and $\hat{\alpha}_g, \hat{\beta}_g, \hat{\gamma}_{ig}$ are estimated using feature-wise ordinary least-squares approach.

Then, given distribution assumptions $Z_{ijg} \sim \mathcal{N}(\gamma_{ig}, \delta_{ig}^2)$, $\gamma_{ig} \sim \mathcal{N}(\gamma_i, \tau_i^2)$, and $\delta_{ig}^2 \sim \text{InverseGamma}(\lambda_i, \theta_i)$, using empirical Bayes algorithm, we can estimate $\gamma_{ig}^*$ and ${\delta_{ig}^*}^2$ iteratively through:
\begin{align}
\label{eq:empirical-bayes}
    \gamma_{ig}^*  = \frac{N_i \overline{\tau}_i^2 \hat{\gamma}_{ig} + {\delta_{ig}^*}^2\overline{\gamma}_i}{N_i \overline{\tau}_i^2 + {\delta_{ig}^*}^2}, \quad\quad
    {\delta_{ig}^*}^2 = \frac{\overline{\theta}_i + \frac{1}{2}\sum_{j}(Z_{ijg} - \gamma_{ig}^*)^2}{\frac{1}{2} N_i + \overline{\lambda}_i - 1},
\end{align}
where $\overline{\tau}_i^2$, $\overline{\gamma}_{i}$, $\overline{\theta_i}$, $\overline{\lambda_i}$ are computed through the method of moments.

Finally, harmonized data is obtained within each site using:
\begin{align}
\label{eq:harmonization}
    y_{ijg}^* = \frac{\hat{\sigma}_g}{\delta_{ig}^*}(Z_{ijg} - \gamma_{ig}^*) + \hat{\alpha}_g + X_{ij}\hat{\beta}_g.
\end{align}

\subsection{\textit{Cluster ComBat}}
\label{sec:cluster-combat}

Though \textsc{ComBat} has been widely adopted for various analyses~\cite{Pomponio2020,Beer2020}, it lacks generalization to new sites. 
When applied to an unseen site, \textsc{ComBat} requires re-estimating all harmonization parameters based on $M+1$ sites, which needs to engage all participating sites to coordinate harmonization, which is costly and usually prohibitive. 
Also, the original \textsc{ComBat} assumes that scale and mean effects exist within each single site, and each group of harmonization parameters (i.e., $\gamma_{ig}$ and $\delta_{ig}$ with the same $i$) can only be estimated within each single site of limited sample size, which may lead to suboptimal estimation of harmonization parameters. 

Instead of assuming harmonization parameters can only be shared within each single site, we assume that multiple sites can share one group of harmonization parameters. 
As such, data points from multiple sites sharing the same harmonization parameters can be clustered into one cluster. 
We thus reformulate the L/S model as:
\begin{align}
    y_{ijg} = \alpha_g + X_{ij}\beta_g + \gamma_{cg} + \delta_{cg}\epsilon_{ijg},
\end{align}
where $c \in [C]$ represents the cluster index of site $i$, and there are $C$ clusters in total, where $C \leq M$. Compared with the original version of ComBat, where we need to estimate $G \cdot M$ harmonization parameters, this cluster-based algorithm only requires the estimation of $G \cdot C$ harmonization parameters.

Using cluster-wise shared harmonization parameters, we can generalize knowledge from previous $N$ sites to the new unseen site once we know which cluster each data point from this site belongs to. Additionally, the estimation process of harmonization parameters $\gamma_{cg}$ and $\delta_{cg}$ can benefit from multiple sites' data points within the same cluster, considering the sample number for estimating each parameter group is enlarged. 
And we name this algorithm \textit{Cluster ComBat}.

The \textit{Cluster ComBat} algorithm requires the following steps for harmonization: 
\begin{inparaenum}[i)]
\quad\item sample clustering using $K$-means, based on data points from all sites, to decide data points' cluster index of each site; 
\quad\item feature-wise standardization on all samples using $\hat{\alpha}_g$, $\hat{\beta}_g$ and $\hat{\sigma}_g$ from least-squares; 
\quad\item empirical Bayes estimation of the cluster-wise harmonization parameters $\gamma_{cg}$ and $\delta_{cg}$ for each cluster based on sites within cluster $c \in [C]$, following Equation~\ref{eq:empirical-bayes} with replacing $N_i$ to the overall sample number in cluster $c$; 
\quad\item harmonization process following Equation~\ref{eq:harmonization} with replacing $\gamma^*_{ig}$ and $\delta^*_{ig}$ to $\gamma^*_{cg}$ and $\delta^*_{cg}$ respectively.
\end{inparaenum} \quad For the cluster index assignment for each training sample, we directly apply sample-wise index assignment using the clustering algorithm, allowing data points at the same site to have different cluster indexes. This is the "privilege" of the centralized setting, as we can access the feature values of all data from all sites, thereby facilitating the determination of the cluster index for each individual data point. This also allows the cluster index of data points to ignore site belonging, considering the reduction of bias not only from sites but also from other potential factors, leading to better handling of data heterogeneity.
The complete process is demonstrated in Algorithm~\ref{alg:ClusterComBat}.

\begin{algorithm}[!t]
    \caption{Centralized \textit{Cluster ComBat}}
    \label{alg:ClusterComBat}
    \begin{algorithmic}
    \Require $y_{ijg}$ - unharmonized data and $X_{ij}$ - biological covariates of sample $j$ from site $i$
    \Ensure $\hat{\alpha}_g, \hat{\beta}_g, \delta_{cg}^*, \gamma_{cg}^*$ - harmonization parameters and $k$ - trained K-means model

    \State Train K-means model $k$ using $y_{ijg}$

    \State Estimate $\hat{\alpha}_{g}, \hat{\beta}_{g},$ and $\hat{\gamma}_{ig}$ using least-square methods

    \State Standardize data via Equation~\ref{eq:standarize} 

    \State Get cluster index $c = k(y_{ij})$ of every $y_{ij}$
    \State Estimate $\delta^*_{cg}, \gamma_{cg}^*$ using $\text{EmpricalBayes}(Z_{ijg})$ via Equation~\ref{eq:empirical-bayes}
    
    \State \Return $\hat{\alpha}_g, \hat{\beta}_g, \delta_{cg}^*, \gamma_{cg}^* , k$
    
    \end{algorithmic}
\end{algorithm}

Now, we introduce how proposed \textit{Cluster ComBat} can apply harmonization to the unseen site $i \notin [M]$. We first use the trained K-means to identify the cluster of each data point $y_{ij}$, denoted as $k(y_{ij})$. Then, with pre-estimated harmonization parameters $\delta^*_{cg}$, $\gamma_{cg}^*$, we can derive the harmonized features for this unseen site $i$ by
\begin{align}
    y_{ijg}^* = \frac{\hat{\sigma}_g}{\delta_{\tilde{c}g}^*}(Z_{ijg} - \gamma_{\tilde{c}g}^*) + \hat{\alpha}_g + X_{ij}\hat{\beta}_g,
\end{align}
where $\tilde{c} = k(y_{ij})$. Algorithm~\ref{alg:ALG4} describes the procedure when dealing with data from an unseen site in the centralized setting.

\begin{algorithm}[!t]
\caption{\textit{Cluster ComBat} for Unseen Site}
    \label{alg:ALG4}
    \begin{algorithmic}
    \Require $y_{ijg}$ - unseen site's data, $X_{ij}$ - biological covariate of new client, previously estimated parameters $\hat{\alpha}_g, \hat{\beta}_g, \delta_{cg}^*, \gamma_{cg}^*$, and trained K-means $k$
    \Ensure $y_{ijg}^*$ - harmonized features

    \State Get cluster index $\tilde{c} = k(y_{ij})$

    \State Standardized data via Equation~\ref{eq:standarize}

    \State \Return $y_{ijg}^* = \frac{\hat{\sigma}_g}{\delta_{\tilde{c}g}^*}(Z_{ijg} - \gamma_{\tilde{c}g}^*) + \hat{\alpha}_g + X_{ij}\hat{\beta}_g$
    
    \end{algorithmic}
\end{algorithm}

\subsection{\textit{Distributed Cluster ComBat}}

In the real-world scenario, large-scale analyses often involve medical data from multiple institutions (e.g., \cite{schijven2023large}). 
The data is often stored in distributed data centers by various data owners, and raw data cannot be transferred to other locations or directly accessed by other institutions (i.e., sites) due to privacy concerns and regulations. 
Thus, centralized algorithms like \textsc{ComBat} cannot be directly applied. Though previous work~\cite{Chen2022} has made an effort to design a distributed version of \textsc{ComBat}, it would face the same problem as \textsc{ComBat} when it comes to the unseen new site.

To this end, we propose \textit{Distributed Cluster ComBat}, extending \textit{Cluster ComBat} by enabling its generalization ability to the unseen site, attributed to the cluster-wise harmonization model. 
However, unlike sample-feature clustering in a centralized setting as shown in Section~\ref{sec:cluster-combat}, we perform clustering on locally estimated feature-wise parameters, e.g., $\alpha_{ig}$, $\beta_{ig}$, and $\gamma_{ig}$, to tackle the inaccessibility of raw samples on other sites. 
The intuition is that if the feature data of sites exhibit a cluster pattern, locally estimated feature-wise parameters will also exhibit the same cluster pattern, which is validated through our simulation studies. Also, the clustering cost is reduced significantly in the distributed version, considering clustering only on $M$ parameter vectors with $M \ll N$.

The \textit{Distributed Cluster ComBat} has the following steps:  
\begin{inparaenum}[i)]
    \item each site estimates feature-wise parameters $\hat{\alpha}_{ig}$, $\hat{\beta}_{ig}$, and $\hat{\gamma}_{ig}$ locally at the same time, and sends parameters to the central server;\quad
    \item the central server performs K-means clustering based on $\hat{\alpha}_{ig}$, $\hat{\beta}_{ig}$, and $\hat{\gamma}_{ig}$ for $i \in [M]$;\quad
    \item the central server aggregates $\{\hat{\alpha}_{ig}\}_{i \in M}$, $\{\hat{\beta}_{ig}\}_{i \in M}$ and $\{\hat{\gamma}_{ig}\}_{i \in M}$ to estimates the global feature-wise parameters $\hat{\alpha}_g$, $\hat{\beta}_g$ and $\hat{\gamma}_{ig}$, and then sends back to each site;\quad
    \item each site standardized the local data using global feature-wise parameters, then locally estimates $\hat{\delta}_{ig}$ and $\hat{\gamma}_{ig}$;\quad
    \item each site sends locally estimated harmonization parameters to the server;\quad
    \item server aggregates harmonization parameters within each cluster to estimate the cluster-wise ones, then sends back to each site; the aggregation procedure precisely follows the procedure outlined in Figure 1 of the original Distributed ComBat algorithm paper~\cite{Chen2022};\quad
    \item each site performs local harmonization based on cluster-wise harmonization parameters.
\end{inparaenum}
The procedure is summarized in Algorithm~\ref{alg:DCComBat}.

When generalized to new unseen site $i \notin [M]$, we first estimate the local feature-wise parameters $\alpha_{ig}$, $\beta_{ig}$ and $\gamma_{ig}$, then use previous trained K-means model $k$ to find the cluster index of the current site based on local estimated feature-wise parameters. 
Others follow a similar procedure as \textit{Cluster ComBat}, as summarized in Algorithm~\ref{alg:ALG2}.

 \begin{algorithm}[!t]
    \caption{\textit{Distributed Cluster ComBat}}
    \label{alg:DCComBat}
    \begin{algorithmic}
    \Require $y_{ijg}$ - unharmonized data and $X_{ij}$ - biological covariates of sample $j$ from site $i$
    \Ensure $\hat{\alpha}_g, \hat{\beta}_g, \delta_{cg}^*, \gamma_{cg}^*$ - Cluster ComBat parameters and $k$ - trained K-means model
    \ForAll{site i}
        \State Estimate $\hat{\alpha}_{ig}, \hat{\beta}_{ig},$ and $\hat{\gamma}_{ig}$ locally using least-squared method from data of site $i$
        \State Send locally estimated $\hat{\alpha}_{ig}, \hat{\beta}_{ig},$ and $\hat{\gamma}_{ig}$ to the central server
    \EndFor

    \State Train K-means model $k$ using $\hat{\alpha}_{ig}, \hat{\beta}_{ig},$ and $\hat{\gamma}_{ig}$

    \State Estimate $\hat{\alpha}_{g}, \hat{\beta}_{g},$ and $\hat{\gamma}_{ig}$ by taking average of all $\hat{\alpha}_{ig}, \hat{\beta}_{ig},$ and $\hat{\gamma}_{ig}$.

    \ForAll{site i}
        \State Standardize local data via \ref{eq:standarize} to get $Z_{ijg}$
        \State Estimate local $\delta_{ig}^*, \gamma_{ig}^*$ using $\text{EmpricalBayes}(Z_{ijg})$ via Equation~\ref{eq:empirical-bayes}
    \EndFor

    \ForAll{cluster c}
        \State Estimate $\delta_{cg}^*, \gamma_{cg}^*$ by taking average of all 
        $\delta_{ig}^*, \gamma_{ig}^*$ for all sites $i$ belong to cluster $c$.
    \EndFor
    
    \State \Return $\hat{\alpha}_g, \hat{\beta}_g, \delta_{cg}^*, \gamma_{cg}^* , k$
    
    \end{algorithmic}
\end{algorithm}
\begin{algorithm}[!ht]
\caption{\textit{Distributed Cluster ComBat} for Unseen Site}
    \label{alg:ALG2}
    \begin{algorithmic}
    \Require $y_{ijg}$ - new client's data, $X_{ij}$ - biological covariate of new client, trained cluster-wise harmonization parameters $\hat{\alpha}_g, \hat{\beta}_g, \delta_{cg}^*, \gamma_{cg}^*$, and trained K-means $k$
    \Ensure $y_{ijg}^*$ - harmonized features
    
    \State Estimate $\hat{\alpha}_{ig}, \hat{\beta}_{ig},$ and $\hat{\gamma}_{ig}$ using least-squared method using data from testing client 

    \State Get cluster index $\tilde{c} = k(\hat{\alpha}_{ig}, \hat{\beta}_{ig},\hat{\gamma}_{ig})$

    \State Standardize data via Equation~\ref{eq:standarize}

    \State \Return $y_{ijg}^* = \frac{\hat{\sigma}_g}{\delta_{\tilde{c}g}^*}(Z_{ijg} - \gamma_{\tilde{c}g}^*) + \hat{\alpha}_g + X_{ij}\hat{\beta}_g$
    \end{algorithmic}
\end{algorithm}

\section{Validation using Simulation}
In simulation, we use controllable synthetic data to validate the correctness of the proposed algorithms and the intuitions used. 

\subsection{Synthetic data generation}
\begin{figure}[h]
  \centering
  \includegraphics[width=\linewidth]{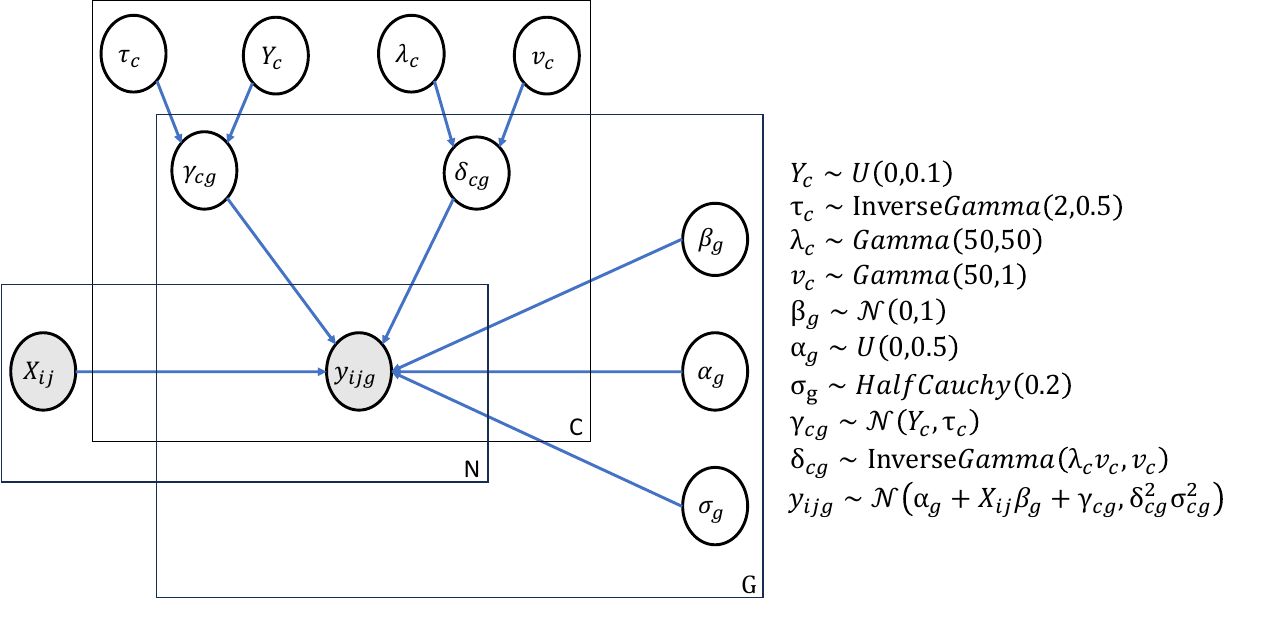}
  \caption{Graphical model used to generate synthetic data. The shaded circles represent observed variables, including biological covariates and feature values, while unshaded circles represent latent parameters.}
  \label{fig:graphical}
\end{figure}

\begin{figure}[!ht]
    \begin{subfigure}{0.5\linewidth}
    \includegraphics[width=\linewidth]{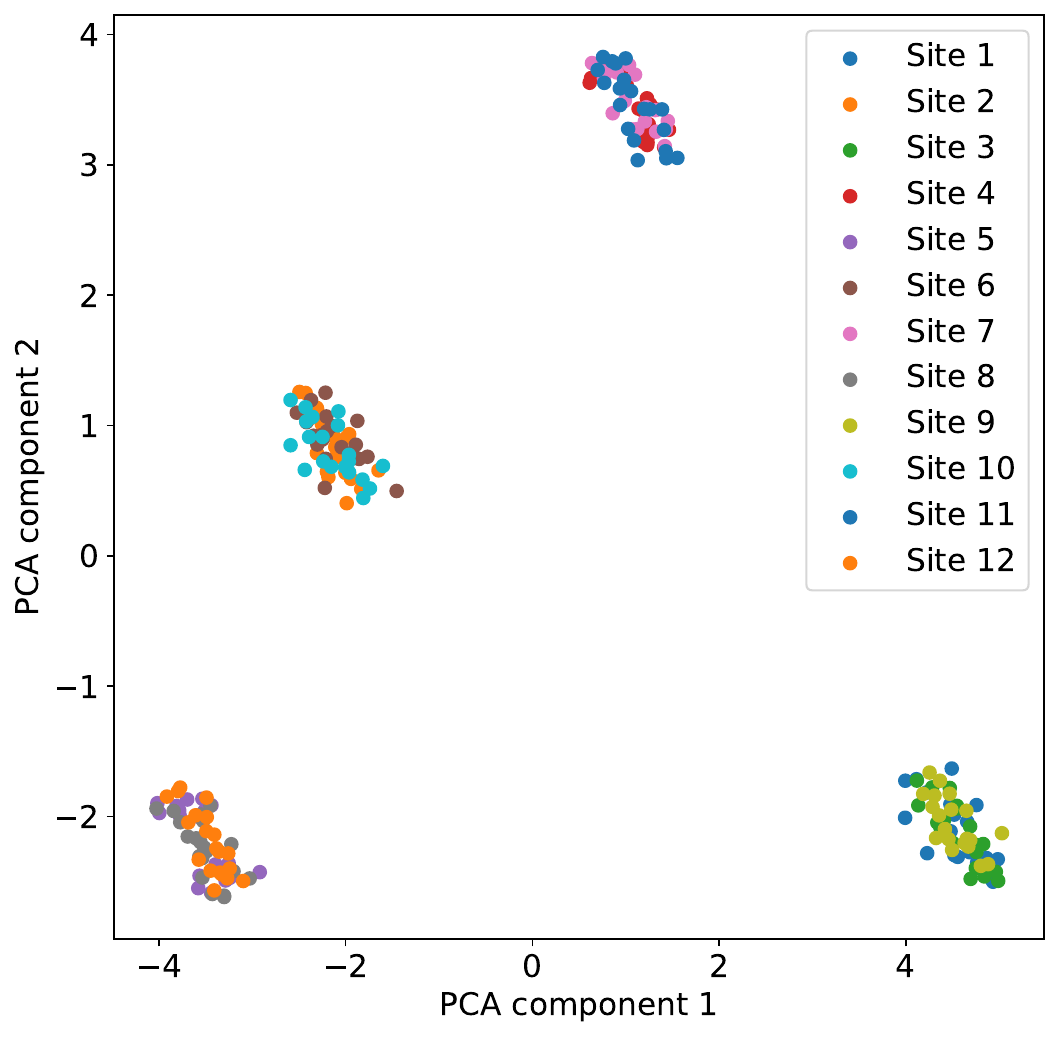}
    \caption{Site pattern}
    \label{fig_site_pattern_raw}
    \end{subfigure}%
    \begin{subfigure}{0.5\linewidth}
    \includegraphics[width=\linewidth]{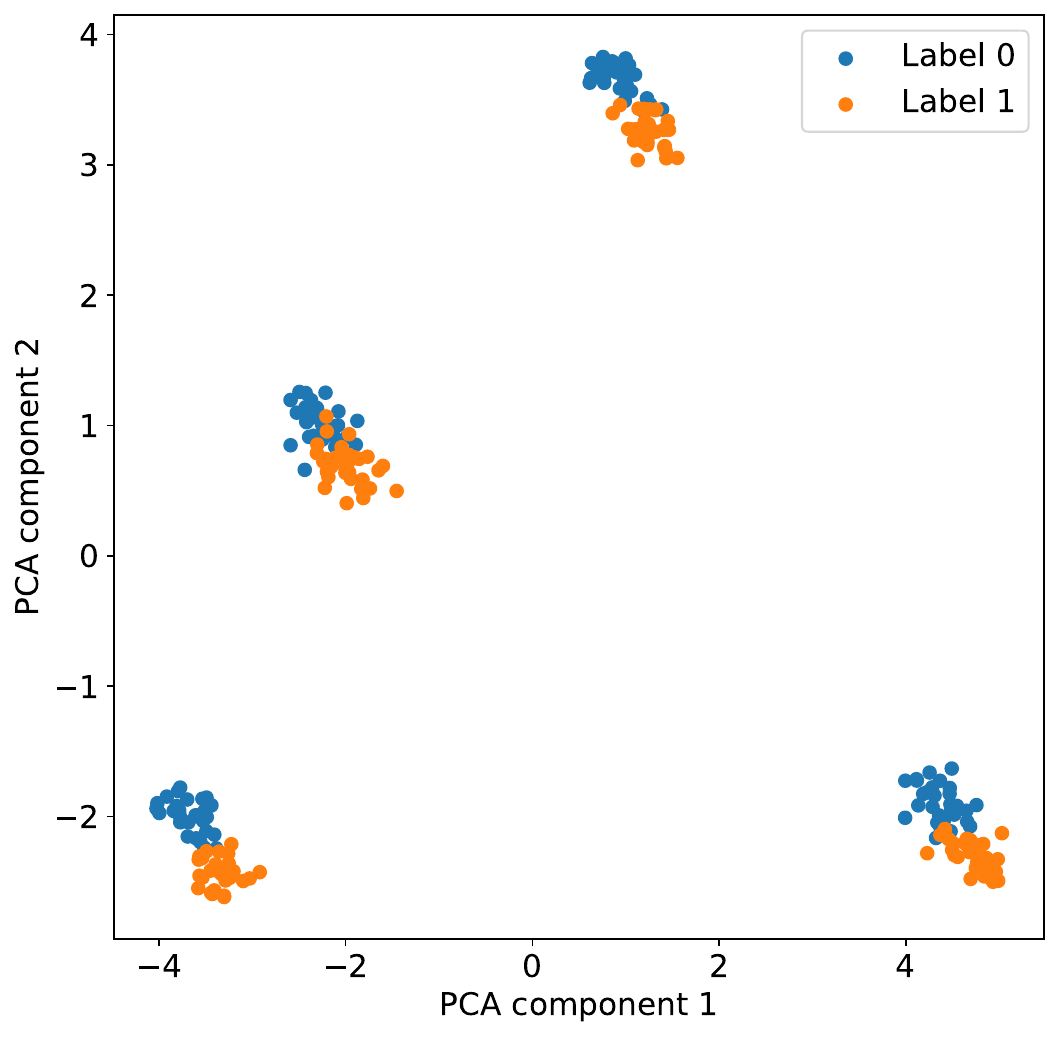}
    \caption{Label pattern}
    \label{fig_label_pattern_raw}
    \end{subfigure}%
\caption{Synthetic Data: site pattern and label pattern of the raw data.}
\end{figure}

We follow the data generation procedure in~\cite{Silva2023} and use the graphical model in Figure \ref{fig:graphical}. 
We replace the site effects with the cluster effects $\gamma_{cg}$.
Specifically, the value $y_{ijg}$ with feature index $g$, site index $i$, and data point index $j$ is considered as  $y_{ijg} \sim \mathcal{N}(\alpha_{g} + X_{ij}\beta_g + \gamma_{cg}, \delta_{cg}^2\sigma_{g}^2)$.
Note that feature values with index $g$ that are from different sites but in the same cluster will be affected by the same cluster effects $\gamma_{cg}$ and $\delta_{cg}$.
The ground truth of the harmonized feature for $y_{ijg}$ (expected feature value after harmonization) is $\alpha_{g} + X_{ij}\beta_g $.
Besides, we induce the task binary label information into the biological covariate $X_{ij}$. For instance, $X_{ij} \sim \mathcal{N}(0.5, 0.5)$ and $X_{ij} \sim \mathcal{N}(-0.5, 0.5)$ imply positive and negative labels, respectively, so that the ground truth data is linearly separated.
To visualize the site pattern, cluster pattern, and label pattern in the synthetic data, Principal Components Analysis (PCA) is employed to reduce the dimension of the data to 2~\cite{Makiewicz1993}. 
Figure~\ref{fig_site_pattern_raw} and Figure~\ref{fig_label_pattern_raw} are examples of visualizing the raw (unharmonized) data, which show the patterns of site, cluster, and downstream task labels.

\subsection{Synthetic data experiment}\label{Simlutation}

\begin{figure}[!ht]
    \begin{subfigure}{0.5\linewidth}
    \includegraphics[width=\linewidth]{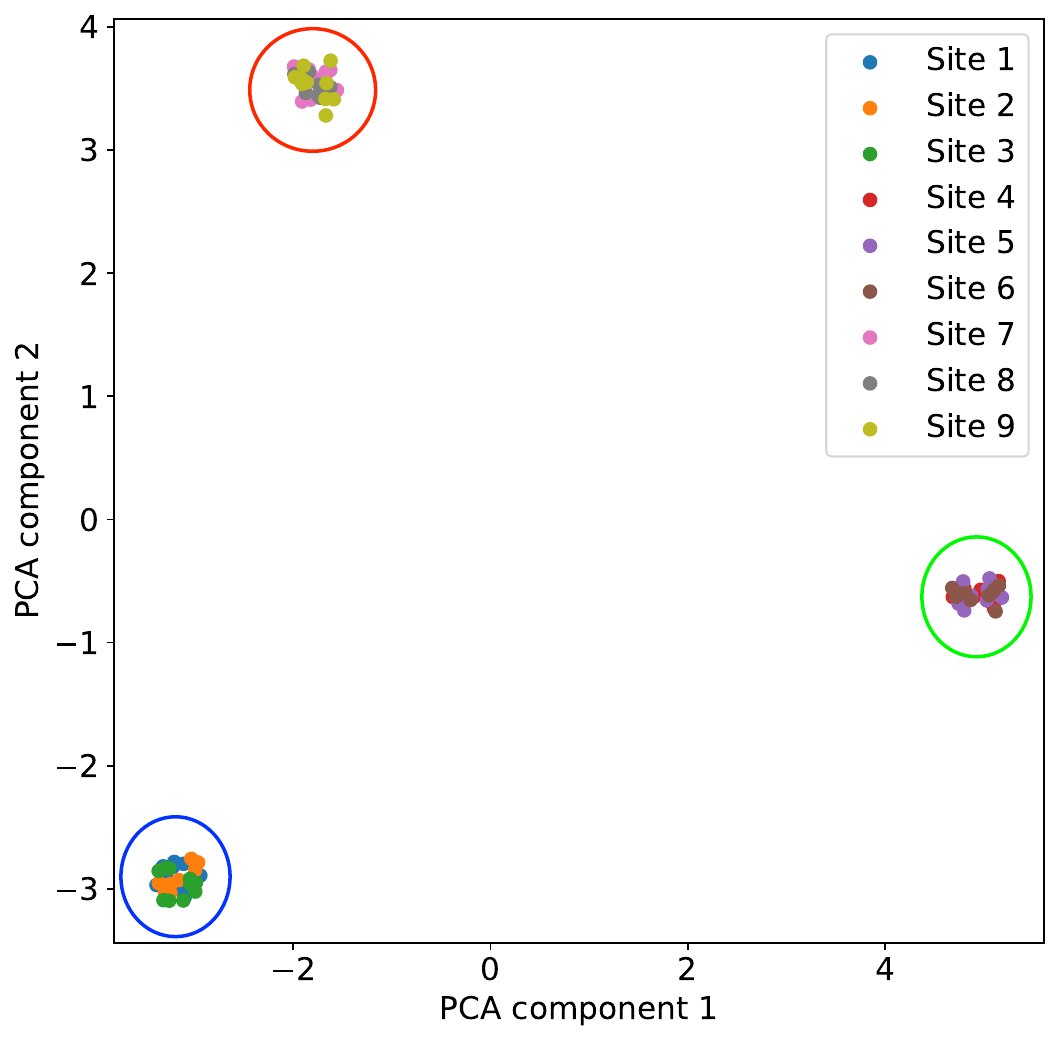}
    \caption{Feature distribution}
    \label{fig:motivation-feature}
    \end{subfigure}%
    \begin{subfigure}{0.5\linewidth}
    \includegraphics[width=\linewidth]{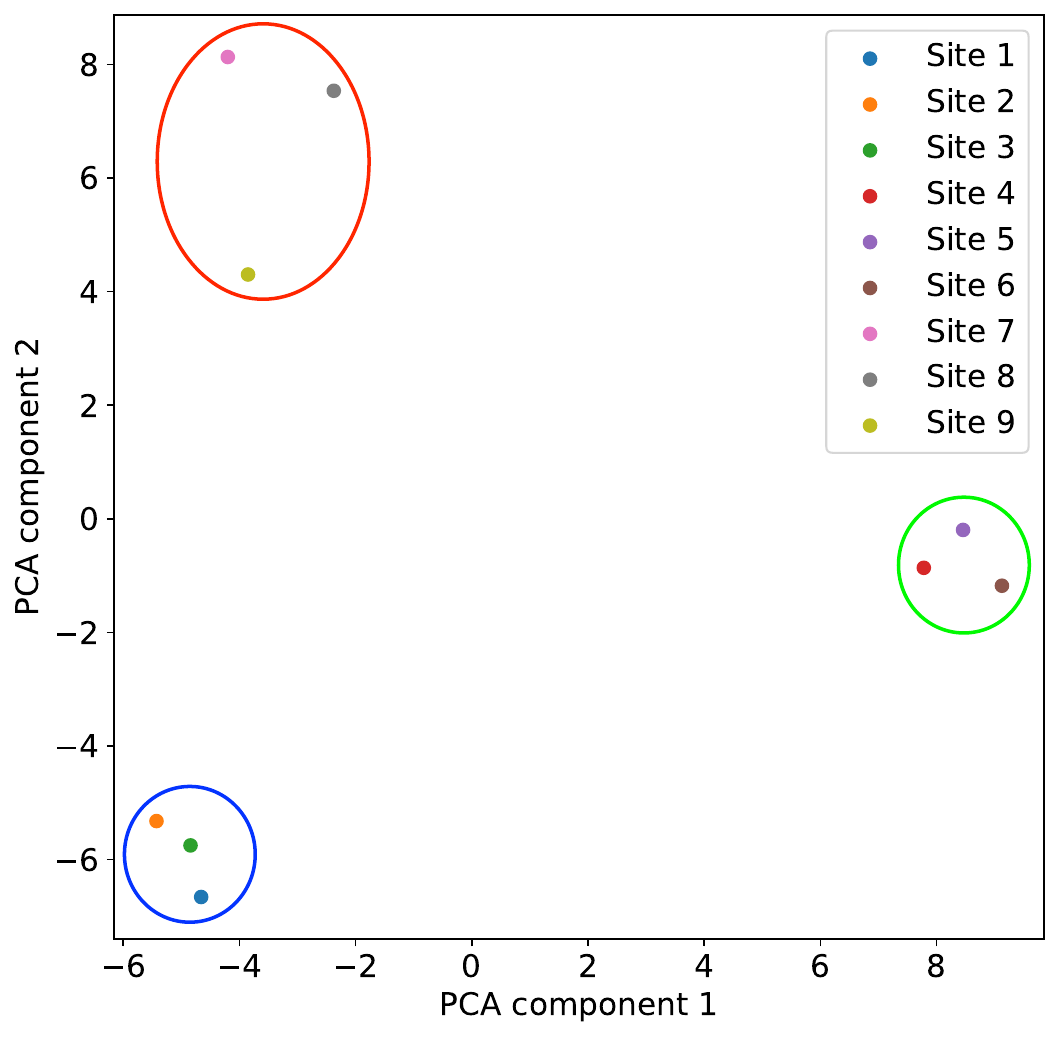}
    \caption{Parameter distribution}
    \label{fig:motivation-param}
    \end{subfigure}%
\caption{Feature and parameter distribution of synthetic data. Sites within the same cluster in the feature space (as shown in (a)) can also be clustered into the same cluster in the parameter space (as shown in (b)). This indicates that cluster patterns in the feature space can be retained in the parameter space.}
\label{fig_parameter}
\end{figure}

We first verify our motivation that if the feature values of sites exhibit cluster patterns in the feature space, then locally estimated feature-wise parameters will also exhibit the same cluster pattern in the parameter space. We generated synthetic data points for 9 sites within 3 clusters for cluster visualization (with data configuration that the number of sites, sample per site, feature, sites per cluster, and biological covariate are 9, 10, 20, 3, and 5 respectively). Specifically, sites 1, 2, and 3 are in the same cluster, sites 4, 5, and 6 are in the same cluster, and sites 7, 8, and 9 are in the same cluster. We used PCA to reduce the dimension to 2 and visualize data points of all sites in the feature space as well as the locally trained parameters for each site in the parameter space. We use colored circles to show the cluster pattern in both feature space and parameter space. As demonstrated in Figure~\ref{fig_parameter}, sites within the same cluster in the feature space (as shown in Figure~\ref{fig:motivation-feature}) can also be clustered into the same cluster in the parameter space (as shown in Figure~\ref{fig:motivation-param}). This indicates that cluster patterns in the feature space can be retained in the parameter space, which verifies our motivation.

\begin{table}[!t]
\small
  \caption{Detailed configurations of synthetic data for simulation}
  \label{tab:information}
  \begin{tabular}{l|ccccc}
    \toprule
    Synthetic Data Index & 1 & 2 & 3 & 4 & 5  \\
    \hline
    $\#\text{Sites}$ & 20 & 25 & 30 & 35 & 40   \\
    $\#\text{Samples Per Site}$ & 20 & 25 & 30 & 35 & 40  \\
    $\#\text{Features}$ & 20 &25 & 30 & 40 & 50  \\
    $\#\text{Sites Per Cluster}$ & 5 & 5 & 5 & 5 & 5 \\
    $\#\text{Biological Covariates}$ & 5 & 5 & 5 & 5 & 5  \\
  \bottomrule
\end{tabular}
\end{table}

Then, we verify the efficacy of our algorithm over \textsc{ComBat} in both centralized and distributed settings using synthetic data. We generate five synthetic data sets, which follow the graphical model Figure~\ref{fig:graphical} with different parameter configurations. 
The detailed generation configurations are summarized in Table~\ref{tab:information}. 
We assess the performance of \textit{Cluster ComBat} harmonization and original \textsc{ComBat} algorithms on the synthetic data with aforementioned conditions over two tasks: ground-truth data, i.e., $\alpha_{g} + X_{ij}\beta_g $, reconstruction task and ground-truth label classification tasks.
Specifically, the Root Mean Square Error (RMSE) between the ground-truth data and the harmonized test data is proposed as the performance measure of the reconstruction task. Also, task accuracy is naturally selected as the performance measure of the downstream classification task.
Because the original \textsc{ComBat} and Distributed ComBat cannot harmonize data from unseen sites, we will retrain harmonization parameters whenever they have data from a testing site. 
Meanwhile, our proposed \textit{Cluster ComBat} and \textit{Distributed Cluster ComBat} can harmonize testing data without the need for retraining the \textsc{ComBat} algorithm. In the experiment, we divided the synthetic data into 70\% for training and 30\% for testing for each task and reported the mean and variance of performance measures over 30 random seeds. 
The results are summarized in Table~\ref{tab:synthetic_reconstruction}.
The results show that \textit{Cluster ComBat} and  \textit{Distributed Cluster ComBat}  outperform \textsc{ComBat} and Distributed ComBat in both tasks over various data conditions.
\begin{table*}[!ht]
\setlength\tabcolsep{0.2em}
\small
\vspace{-0.1in}
  \caption{Validate proposed method using synthetic data. $^{[a]}$ means retraining with test sites.}
  \vspace{-0.1in}
  \label{tab:synthetic_reconstruction}
  \centering
  \begin{threeparttable}
  \begin{tabular}{c|ccccc|ccccc}
    \toprule
    \multirow{3}{*}{Algorithm} & \multicolumn{10}{c}{Performance over different synthetic data conditions}\\
     & \multicolumn{5}{c}{Data Reconstruction Task (RMSE)} & \multicolumn{5}{c}{Data Downstream Task (Acc.)}\\
     \cmidrule(l{4pt}r{4pt}){2-6}
    \cmidrule(l{4pt}r{4pt}){7-11}
     & Data-1 &  Data-2 &  Data-3 &  Data-4 &  Data-5& Data-1 &  Data-2 &  Data-3 &  Data-4 &  Data-5\\
    \midrule
    \multicolumn{11}{l}{Centralized Setting}  \\
     \midrule
     Without harmonization & 14.35±0.28 &35.65±12.15 & 22.07±1.07 & 31.53±9.87 &  31.67±3.03 & 96.97±0.02 & 97.57±0.01 & 98.42±0.00 &  98.59±0.00 & 98.85±0.00  \\
     \textsc{ComBat}$^{[a]}$ & 6.53±0.03 & 14.49±1.73 &7.40±0.08 &11.74±0.90 & 9.16±0.23  & 96.92±0.02 & 90.10±0.14 & 98.57±0.00 & 98.65±0.00 & \textbf{99.00±0.00} \\
     \textit{Cluster ComBat} & \textbf{6.43±0.05} & \textbf{14.38±2.75} & \textbf{7.29±0.12} & \textbf{11.70±1.22} & \textbf{9.06±0.32} & \textbf{97.03±0.01} & \textbf{97.93±0.00}& \textbf{98.59±0.00} & \textbf{98.84±0.00} & 98.94±0.00\\  
     
    \midrule
     \multicolumn{11}{l}{Decentralized Setting} \\
     \midrule
     Distributed ComBat$^{[a]}$ &  6.60±0.03&14.54±1.58& 7.42±0.07& 11.75±0.84& 9.19±0.21 & 95.53±0.04& 90.65±0.27& 97.47±0.01& 98.02±0.01& 97.35±0.03  \\
      \textit{Distributed Cluster ComBat} & \textbf{6.44±0.05}& \textbf{14.37±2.62}& \textbf{7.28±0.11}& \textbf{11.69±1.17}& \textbf{9.04±0.31} & \textbf{97.22±0.02}& \textbf{97.70±0.01}& \textbf{98.68±0.00}&\textbf{98.77±0.00}& \textbf{98.93±0.00}\\  
     
  \bottomrule
\end{tabular}
\vspace{-0.1in}
    \end{threeparttable}
\end{table*}

\begin{figure}[!ht]
    \begin{subfigure}{0.33\linewidth}
    \includegraphics[width=\linewidth]{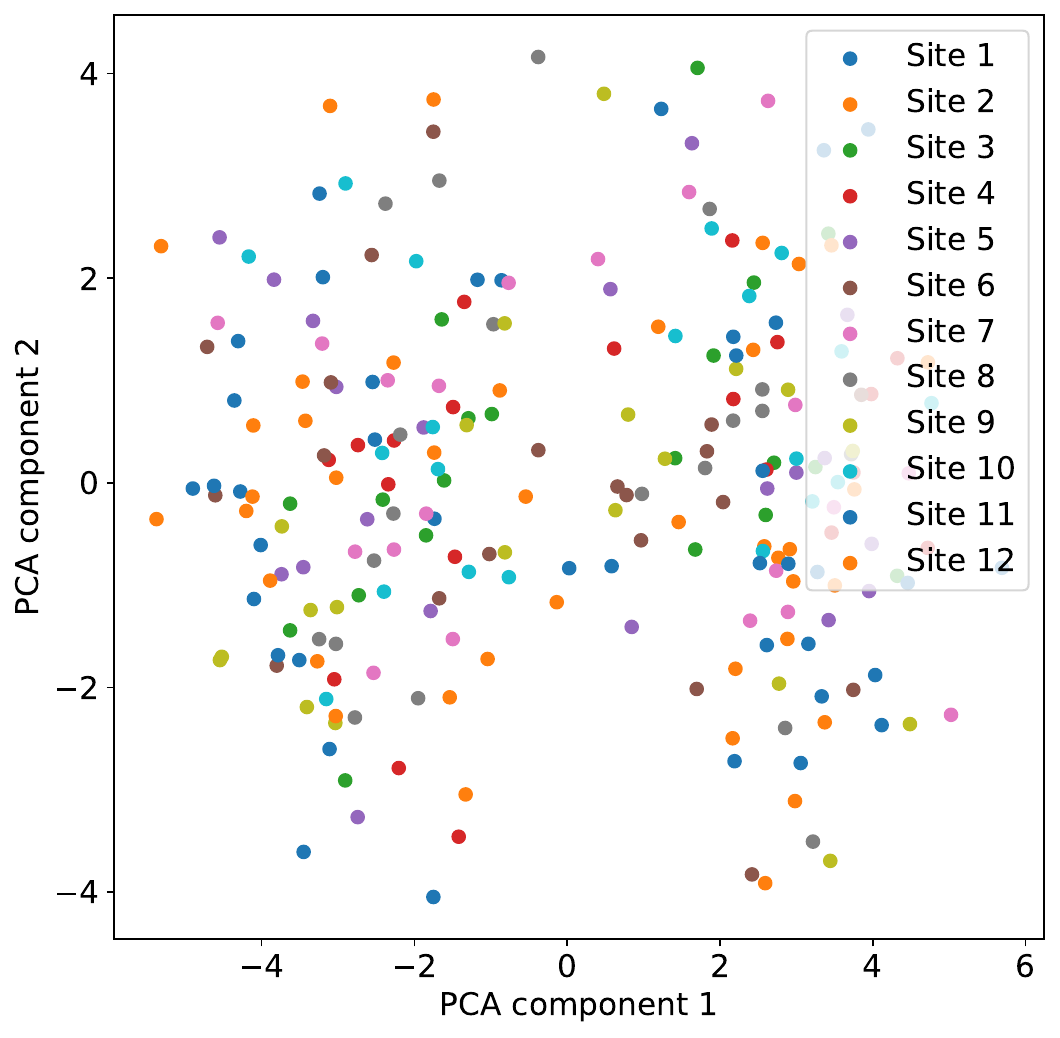}
    \caption{Ground Truth}
    \label{fig_site_raw}
    \end{subfigure}%
    \begin{subfigure}{0.33\linewidth}
    \includegraphics[width=\linewidth]{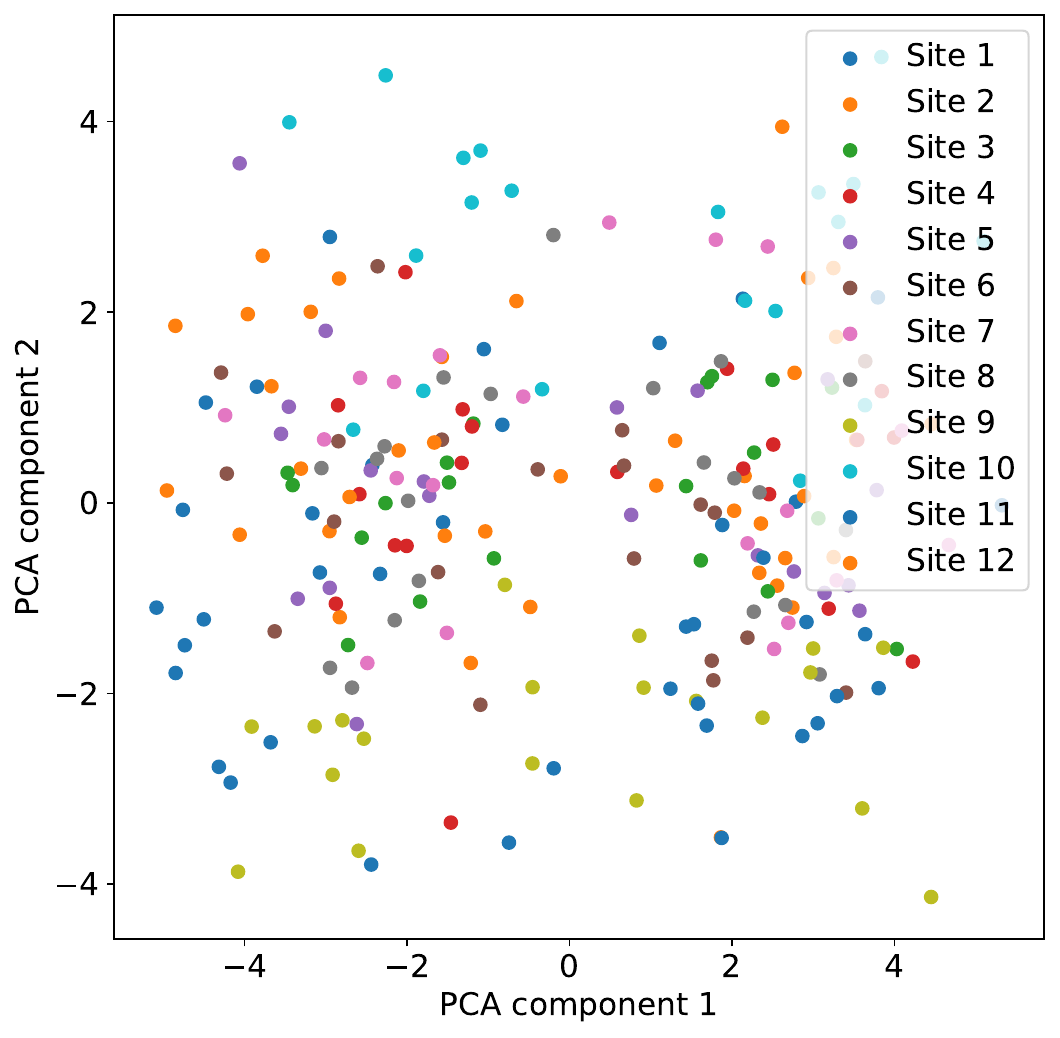}
    \caption{\textsc{ComBat}}
    \label{fig_site_combat}
    \end{subfigure}%
    \begin{subfigure}{0.33\linewidth}
    \includegraphics[width=\linewidth]{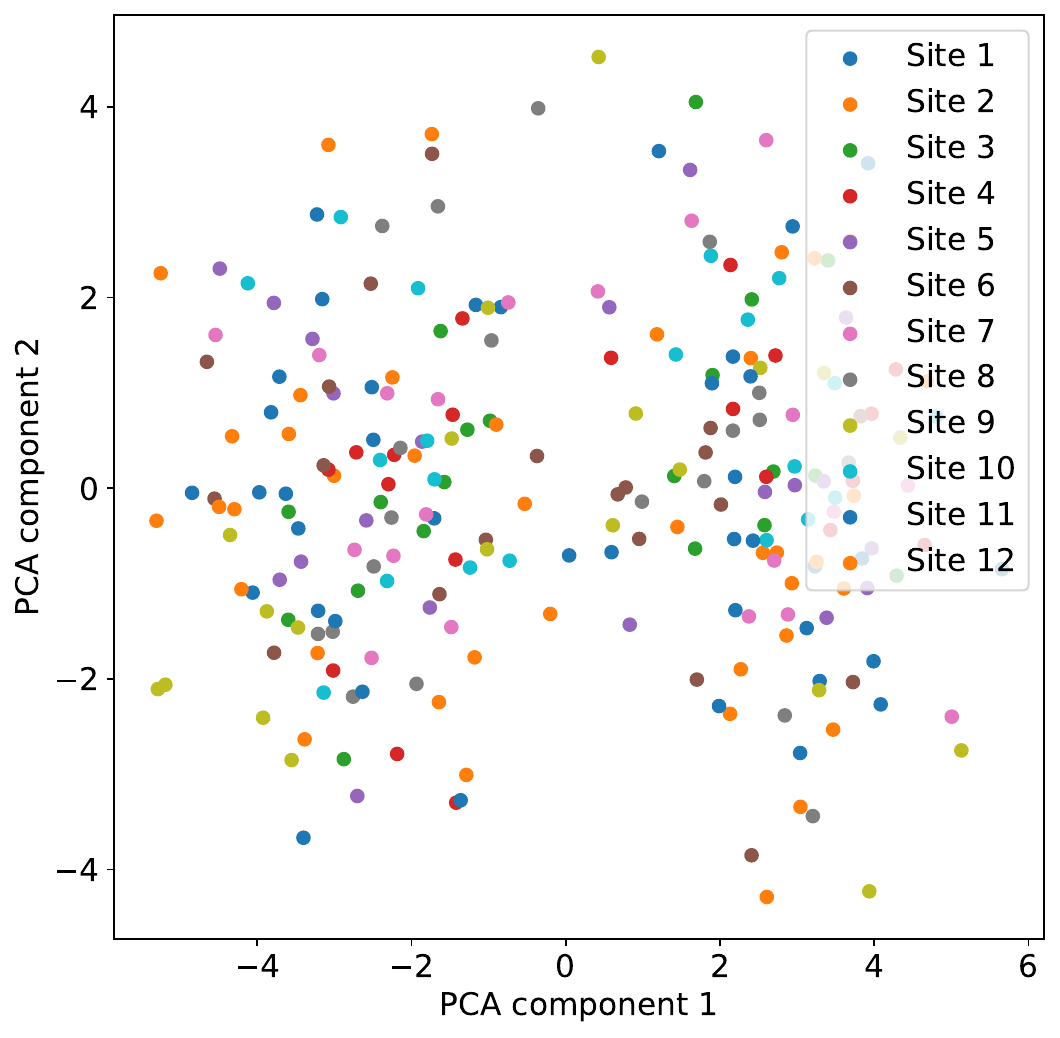}
    \caption{\textit{Cluster ComBat}}
    \label{fig_site_clusterconbat}
    \end{subfigure}%

    \begin{subfigure}{0.33\linewidth}
    \includegraphics[width=\linewidth]{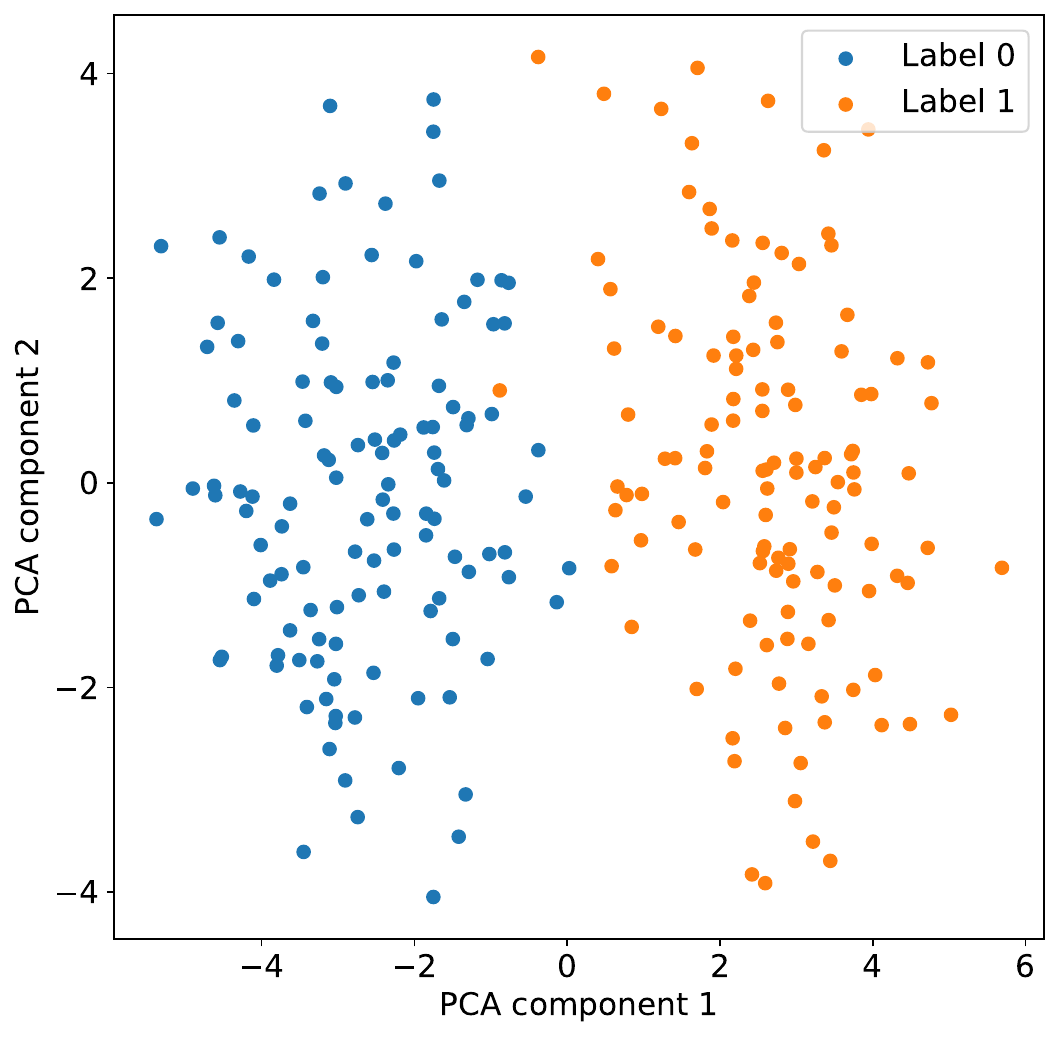}
    \caption{Ground Truth}
    \label{fig_label_raw}
    \end{subfigure}%
    \begin{subfigure}{0.33\linewidth}
    \includegraphics[width=\linewidth]{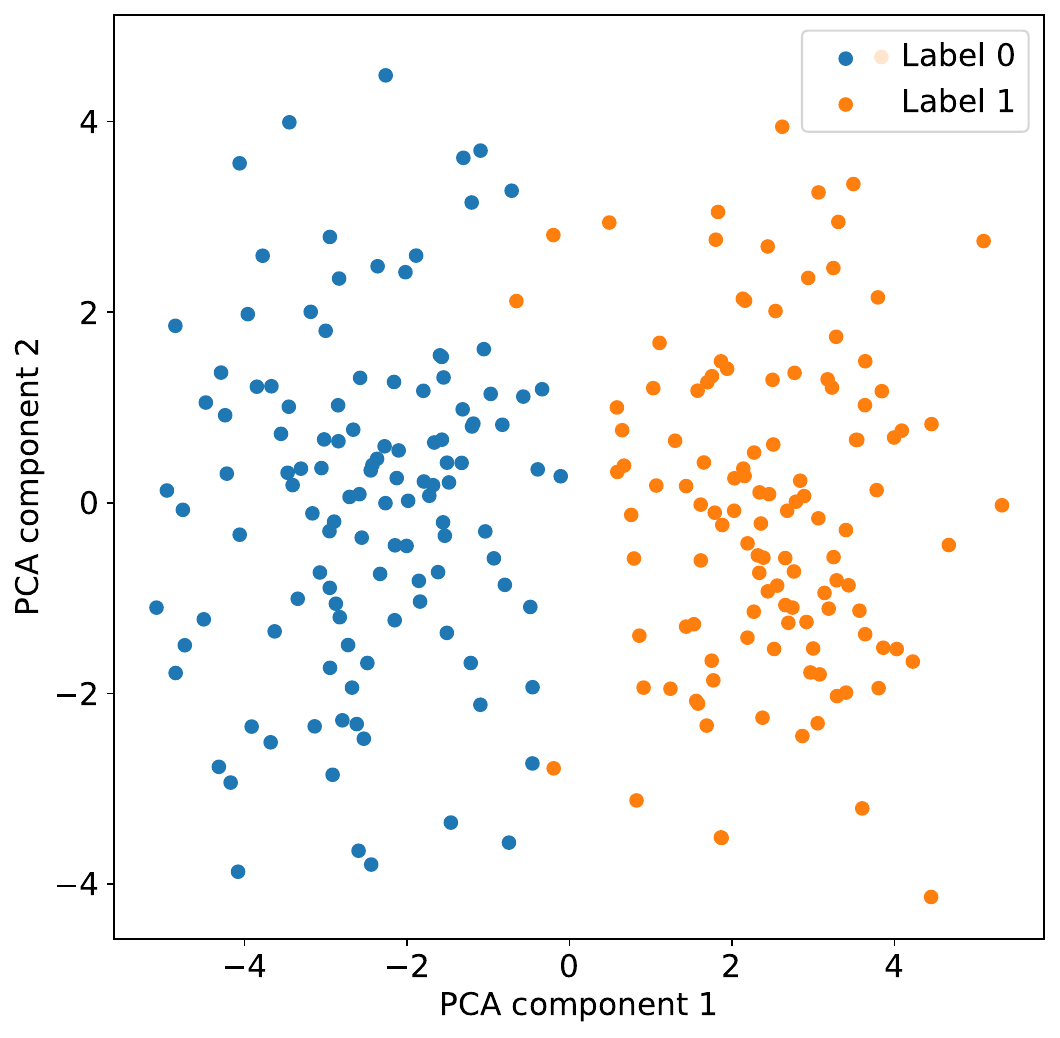}
    \caption{\textsc{ComBat}}
    \label{fig_label_combat}
    \end{subfigure}%
    \begin{subfigure}{0.33\linewidth}
    \includegraphics[width=\linewidth]{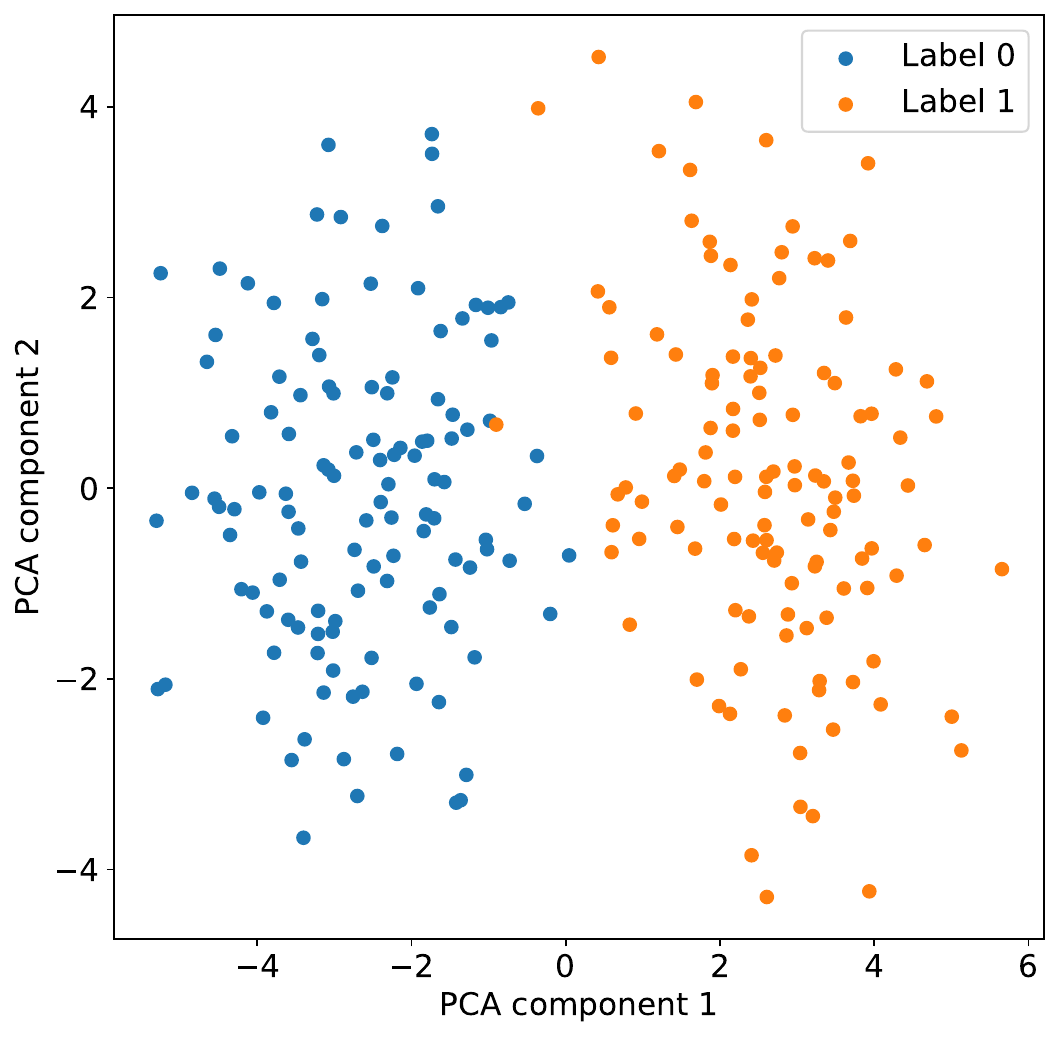}
    \caption{\textit{Cluster ComBat}}
    \label{fig_label_ClusterComBat}
    \end{subfigure}%
\caption{Synthetic Data: site pattern (Figure~\ref{fig_site_raw}, \ref{fig_site_combat}, \ref{fig_site_clusterconbat}) and label pattern (Figure~\ref{fig_label_raw}, \ref{fig_label_combat}, \ref{fig_label_ClusterComBat}) after harmonization.}
\label{fig_partten_vis}
\end{figure}

Besides, Figure \ref{fig_partten_vis} is an example of demonstrating the site pattern and cluster pattern after harmonization (with data configuration that the number of sites, sample per site, feature, sites per cluster, and biological covariate are 12, 20, 20, 3 and 10 respectively).
We see that both harmonization methods maintain the task label information, but the site information has been largely erased.
We want to reiterate that the original \textsc{ComBat}, both in centralized or decentralized settings, requires the retraining procedure when harmonizing the testing data. 
On the contrary, the proposed \textit{Cluster ComBat} eliminates the requirement, benefiting from our parameter-free cluster procedure on unseen data.

\section{Validation on Brain Imaging}

\subsection{ADNI Data}
We use neuroimaging data from the second phase of the North American Alzheimer's Disease Neuroimaging Initiative (ADNI) to evaluate our proposed methods. 
The ADNI data we used has MRI imaging of 563 scans/subjects collected from 18 participating sites. 
We extracted regional measures from DTI data, following the procedure in~\cite{nir2013effectiveness}, leading to 228 features from each scan. 
We also construct a set of downstream prediction tasks, including the prediction of a set of ADNI-defined indicators derived from the neuropsychological battery to characterize memory, executive function, and language. Specifically: 
1) MEM: The ADNI-Mem composite score for memory, which is based on the Rey Auditory Verbal Learning task, word list learning and recognition tasks from ADAS-Cog, recall from Logical Memory I of the Wechsler Memory Test–Revised, and the 3-word recall item from the MMSE~\cite{crane2012development}. 
2) EXF: ADNI-EF composite score for executive function, including Category Fluency (i.e., animals and vegetables), Trail-Making Test parts A and B, Digit Span Backwards, Wechsler Adult Intelligence Scale–Revised Digit–Symbol Substitution, and 5 Clock Drawing items~\cite{gibbons2012composite}. 
3) LAN: ADNI-Lan indicator, which is a composite measure of language~\cite{gibbons2012composite}. 
We also include changes in these scores from baselines~\cite{hohman2017evaluating}, denoted by MEM SLOPES, EXF SLOPES, and LAN SLOPES, respectively. Later, we use these six target variables to evaluate regression performance in downstream tasks. 
The characteristic distribution of the ADNI dataset is illustrated in Table~\ref{tab:characteristic}. 

\begin{table}[!t]

\small
\setlength\tabcolsep{0.2em}
  \caption{Characteristic Distribution of ADNI dataset}
  \vspace{-0.1in}
  \label{tab:characteristic}
  \centering
  \begin{tabular}{c|cccc}
  
    \toprule
     Variable & All (n = 563) & NL (n = 178) & MCI (n = 292) & AD (n = 93)  \\
    \midrule
    Age & 75.06±7.28 & 75.72±6.70 & 74.44±7.40 & 75.74±7.81 \\
    Gender (\%women) & 41.39 & 44.94 & 41.44 & 34.41 \\
  
   $\#\text{Samples Per Site}$ & 31.28±19.28 & 9.89±9.38 & 16.22±10.50 & 5.17±5.96 \\
    MEM & 0.24±0.74 & 0.86±0.53 & 0.22±0.45 & -0.85±0.45 \\
    MEM SLOPES & -0.09±0.11 &  -0.04±0.06 & -0.07±0.08 & -0.25±0.09\\
    EXF & 0.45±0.62 & 0.78±0.51 & 0.47±0.47 & -0.25±0.66\\
    EXF SLOPES & -0.06±0.08 & -0.03±0.05 & -0.05±0.08 & -0.13±0.08 \\
    LAN & 0.45±0.67& 0.85±0.44 & 0.43±0.53 &  -0.26±0.79 \\
    LAN SLOPES & -0.07±0.09&-0.03±0.05&-0.06±0.07& -0.18±0.10\\
  \bottomrule
  
\end{tabular}

\vspace{-0.1in}
\end{table}

\subsection{Site and Cluster Effects in Brain Imaging}

We first show that site effect and cluster effect do exist in ADNI imaging data. We perform two classification tasks on brain imaging:
\begin{inparaenum}[i)]
\item site classification, and \item cluster classification.
\end{inparaenum}
For both tasks, the inputs are the raw feature values of the brain imaging samples, and the output labels are the site index for the site classification task and the cluster index for the cluster classification task. 
We show that harmonization (both \textsc{ComBat} and \textit{Cluster ComBat}) makes it difficult to distinguish samples from different sites/clusters, i.e., lower site/cluster classification accuracy after harmonization.

For site classification, the site index is a sample's natural site index, and the overall class number is 18. 
For cluster classification, we perform K-means to cluster 18 sites into 5 clusters to assign cluster indexes, and thus the overall class number is 5. 
Specifically, samples with the same cluster indexes can come from the same site or different sites, while samples with different cluster indexes must come from different sites. 
Logistic Regression is used for both tasks to classify brain imaging.
For the train/test split of both tasks, we randomly select $70\%$ of brain imaging as the training set and the remaining $30\%$ for the testing set. 
The test accuracy results of both tasks are averaged over 100 runs with different random seeds.

Table~\ref{tab:siteClass} shows that logistic regression achieves high test accuracy on unharmonized DTI imaging for both tasks. 
By applying either \textsc{ComBat} or \textit{Cluster ComBat} harmonization, the test accuracy drops significantly, indicating that either harmonization method makes it harder for the classifier to distinguish between different sites/clusters. 
This shows that both site/cluster effects on real brain imaging and harmonization methods can alleviate these effects. 
We also notice that \textit{Cluster ComBat} has higher accuracy compared with \textsc{ComBat} in site classification with similar accuracy in cluster classification. 
This can be explained as that after removing cluster effect based on cluster-wise harmonization parameters, differences between clusters are removed by \textit{Cluster ComBat}, while site differences still exist among sites within the same cluster. 
Thus, it is still possible to differentiate between sites within each cluster even after harmonization in \textit{Cluster ComBat} case. 
This shows that our assumption for cluster-wise harmonization works well on real brain imaging data. 

\begin{table}[!t]
\small
  \caption{Accuracy of site and cluster classification on brain imaging data}
  \label{tab:siteClass}
  \begin{tabular}{lcc}
    \toprule
    Harmonization Algorithm & Site & Cluster  \\
    \midrule
    Without harmonization & 86.98±0.078 &  82.23±0.067\\
    \textit{Cluster ComBat} &  36.70±0.091 & 19.32±0.073 \\
    \textsc{ComBat} &  6.93±0.034 & 20.75±0.076 \\
  \bottomrule
\end{tabular}
\end{table}

\begin{figure}[!t]
\begin{subfigure}{.33\linewidth}
  \centering
  \includegraphics[width=\linewidth]{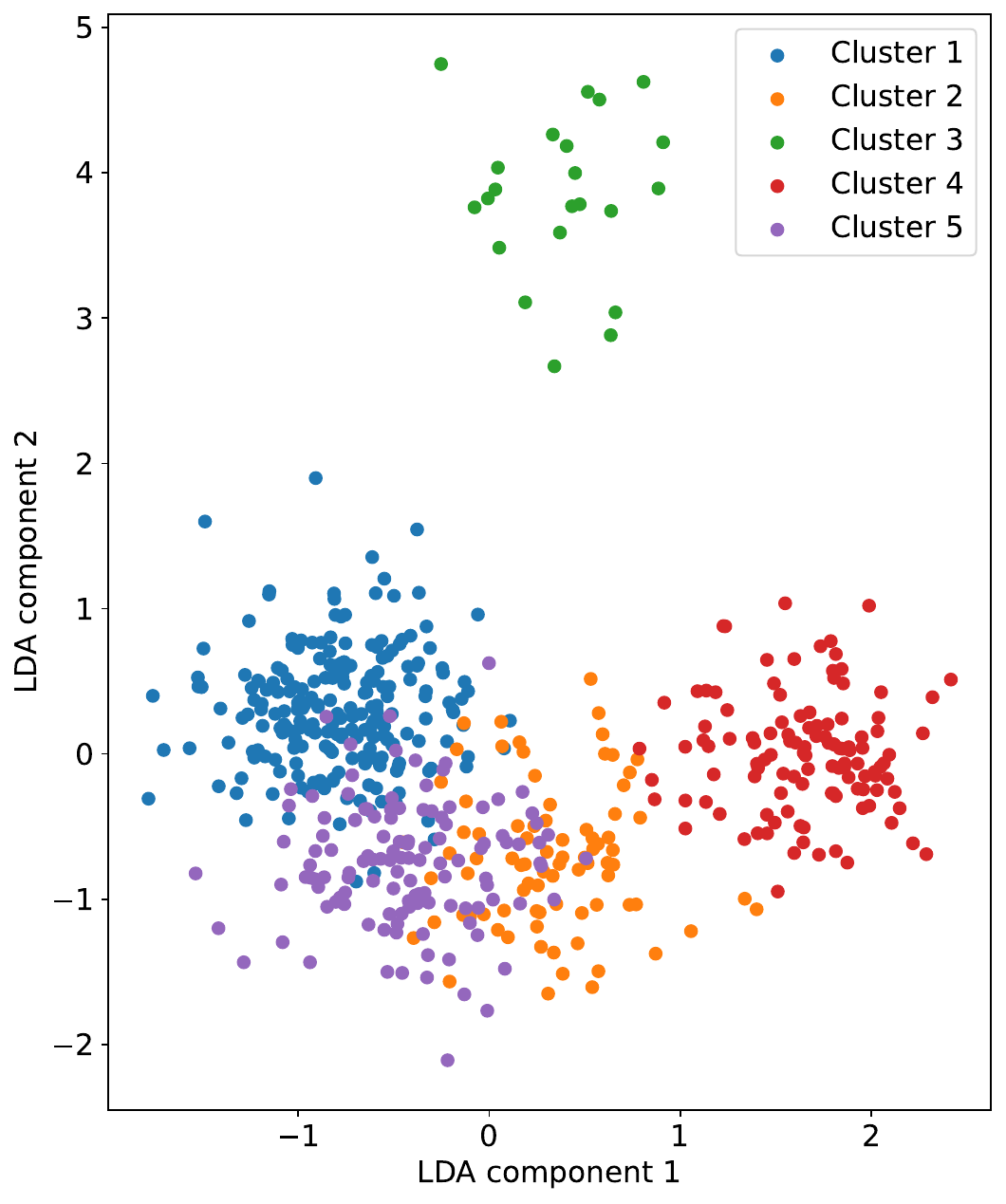}
  \caption{Unharmonized}
  \label{fig:ClusterEffect-unharm}
\end{subfigure}%
\begin{subfigure}{.33\linewidth}
  \centering
  \includegraphics[width=\linewidth]{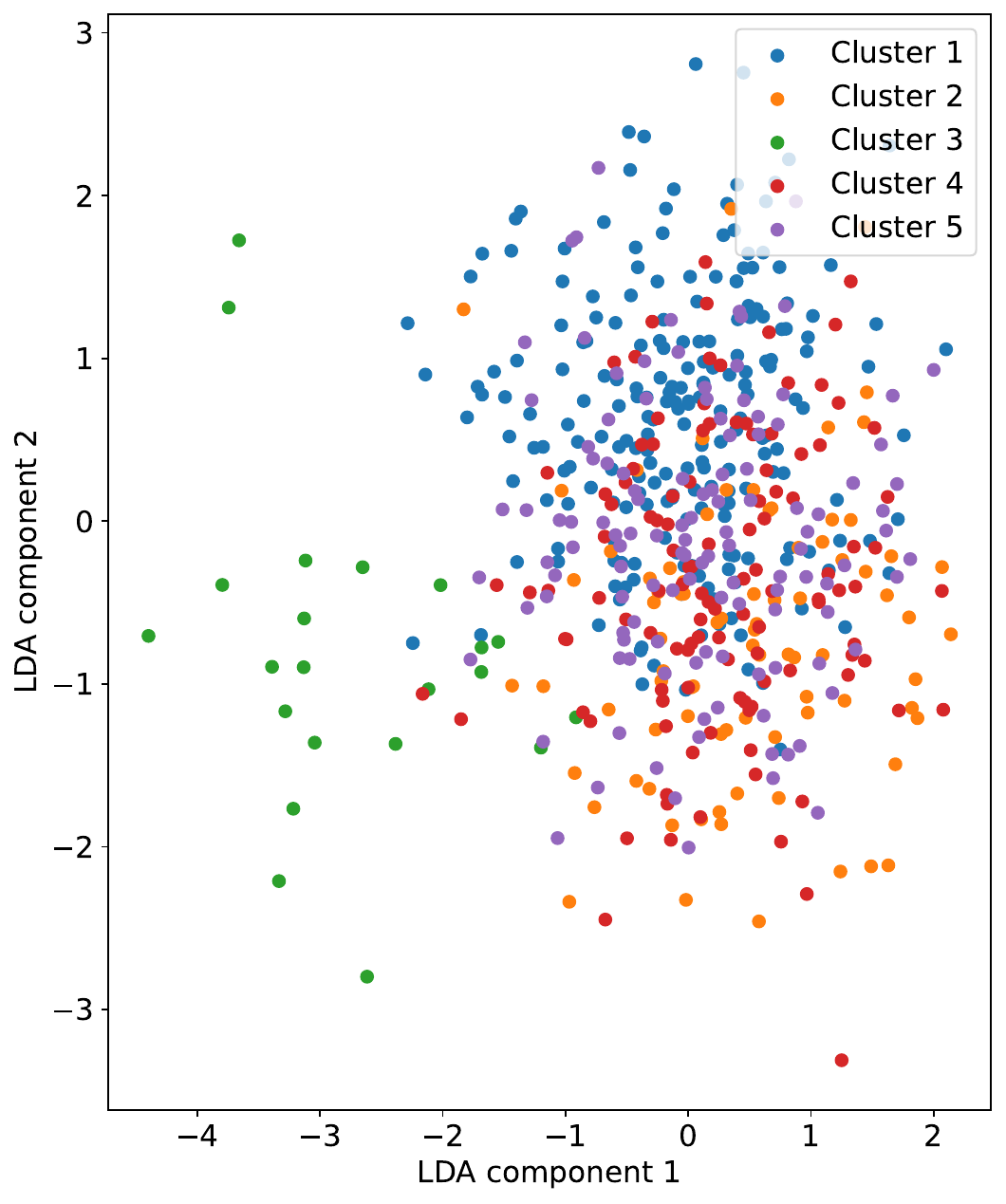}
  \caption{\textsc{ComBat}}
  \label{fig:ClusterEffect-combat}
\end{subfigure}%
\begin{subfigure}{.33\linewidth}
  \centering
  \includegraphics[width=\linewidth]{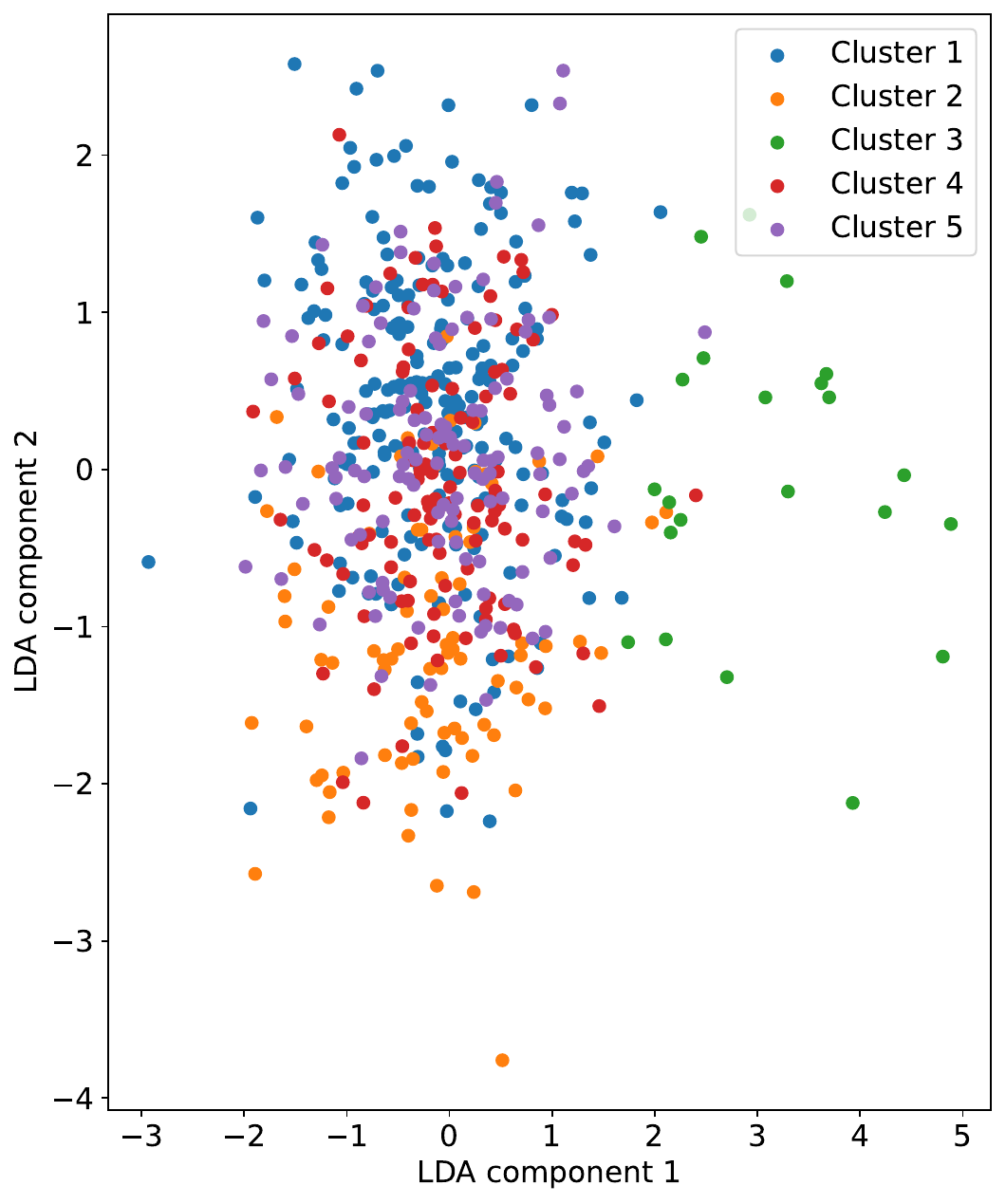}
  \caption{\textit{Cluster ComBat}}
  \label{fig:ClusterEffect-cluster-combat}
\end{subfigure}%
\vspace{-0.1in}
\caption{LDA plot of brain imaging data by cluster index}
\vspace{-0.1in}
\label{fig:ClusterEffect}
\end{figure}

\begin{figure}[!t]
\begin{subfigure}{.5\linewidth}
  \centering
  \includegraphics[width=\linewidth]{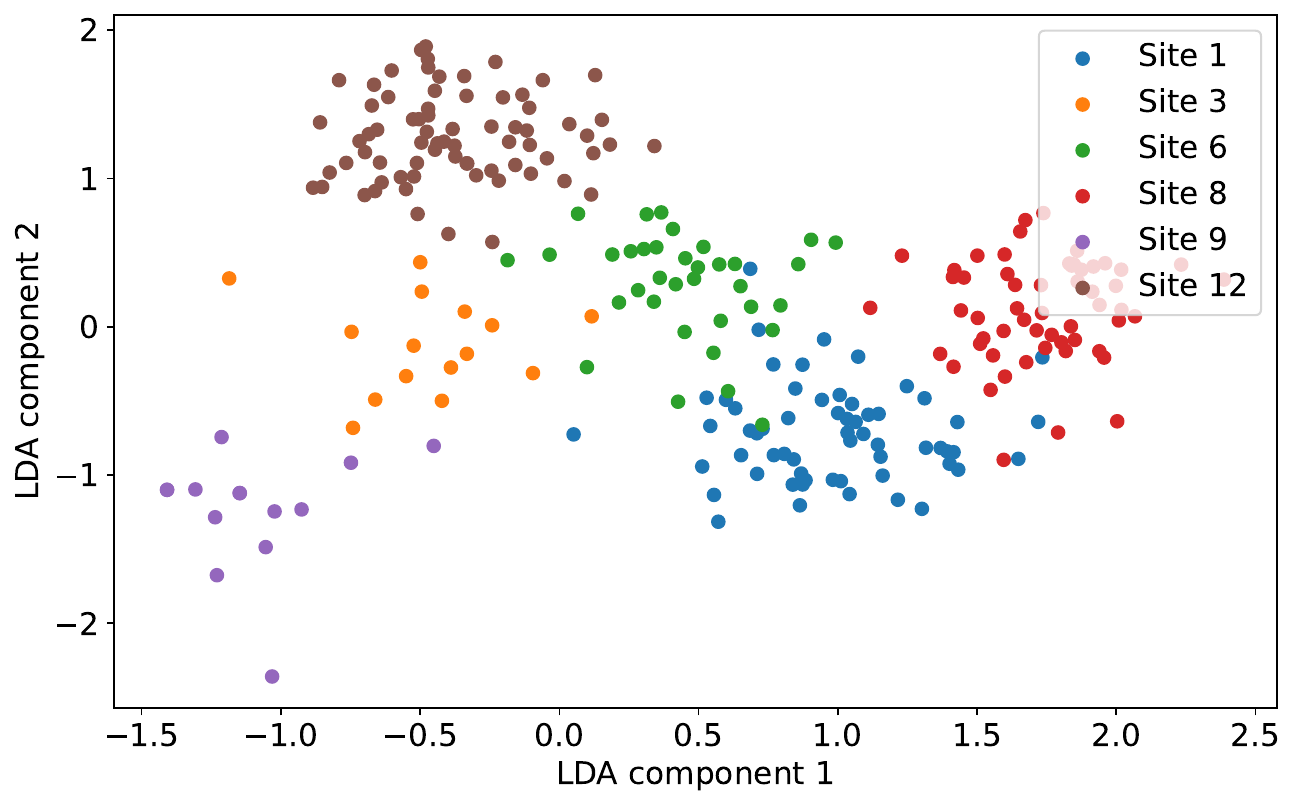}
  \caption{Unharmonized$^{[b]}$}
  \label{fig:SiteEffect-unharm-sitecolor}
\end{subfigure}%
\begin{subfigure}{.5\linewidth}
  \centering
  \includegraphics[width=\linewidth]{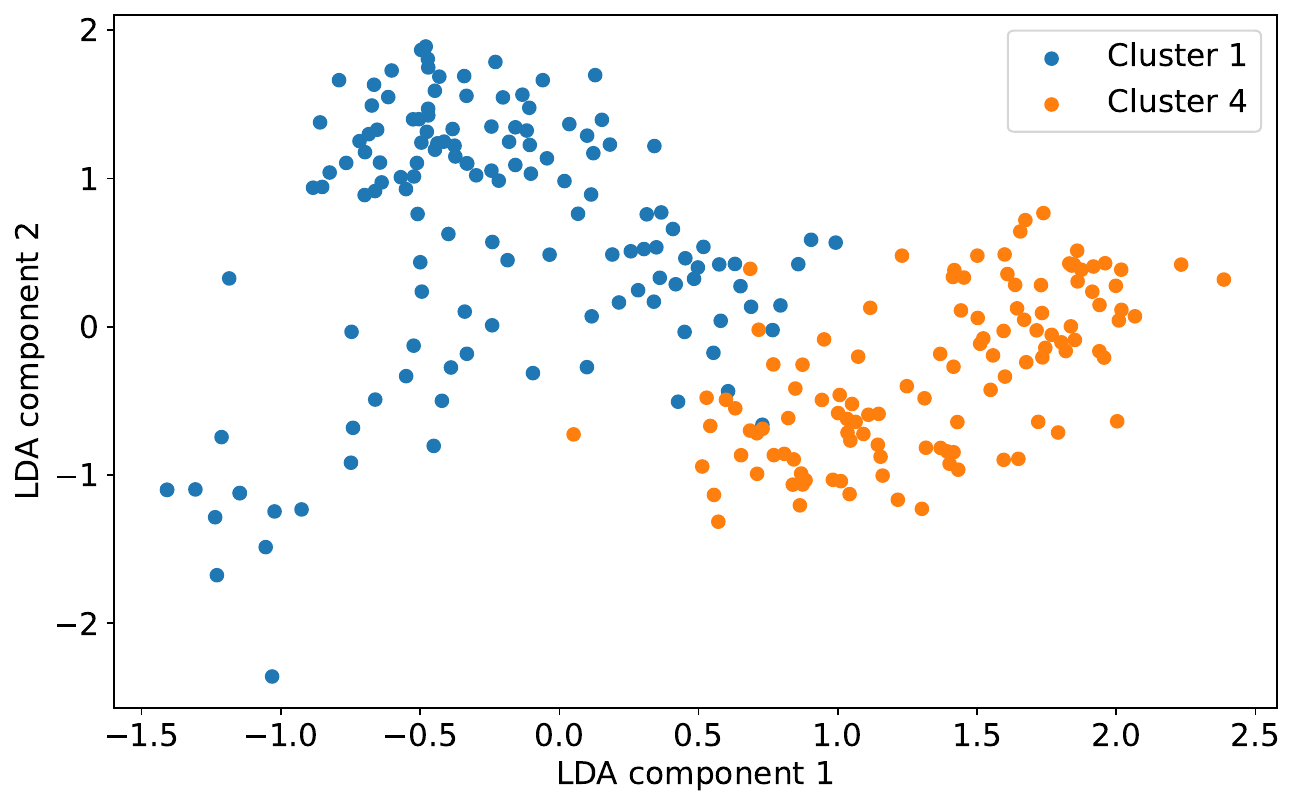}
  \caption{Unharmonized$^{[c]}$}
  \label{fig:SiteEffect-unharm-clustercolor}
\end{subfigure}%

\begin{subfigure}{.5\linewidth}
  \centering
 \includegraphics[width=\linewidth]{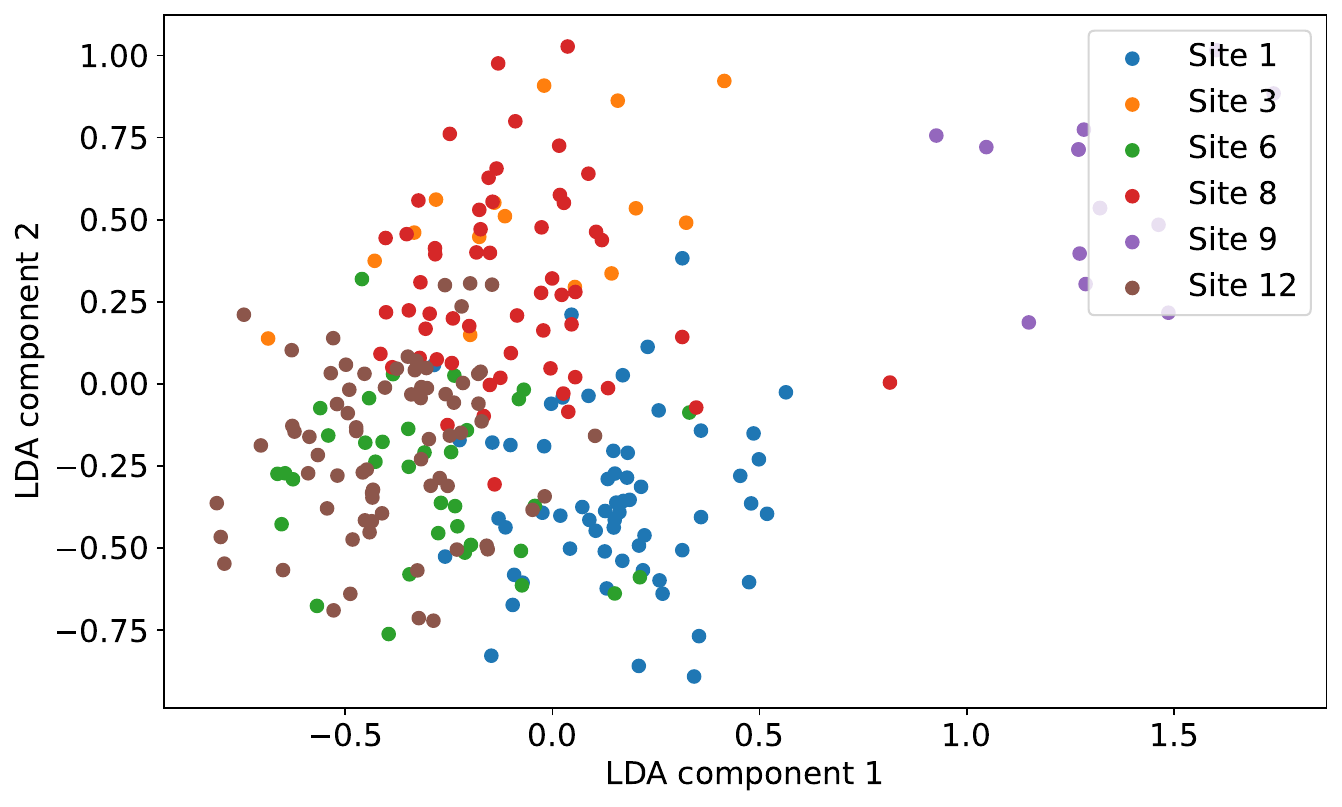}
  \caption{\textit{Cluster ComBat}$^{[b]}$}
  \label{fig:SiteEffect-clustercombat-sitecolor}
\end{subfigure}%
\begin{subfigure}{.5\linewidth}
  \centering
  \includegraphics[width=\linewidth]{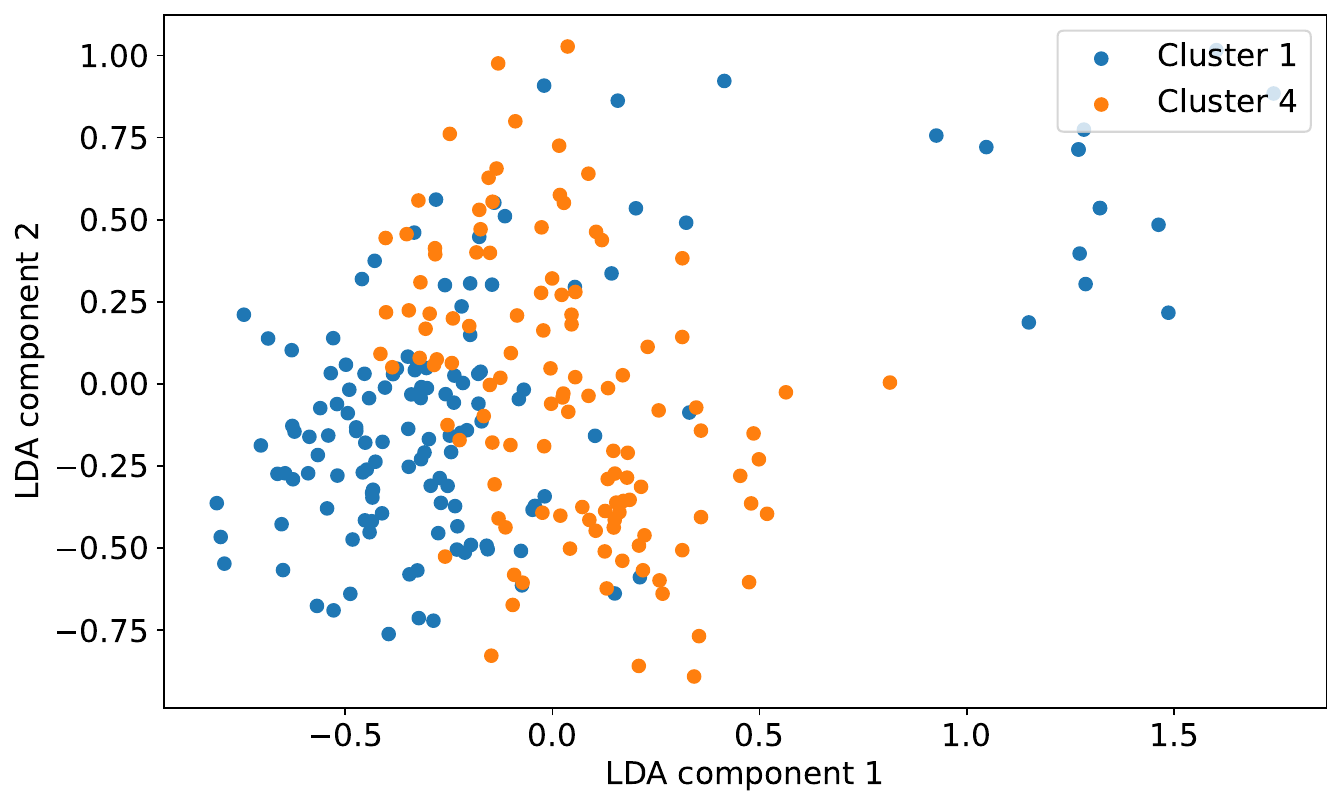}
  \caption{\textit{Cluster ComBat}$^{[c]}$}
  \label{fig:SiteEffect-clustercombat-clustercolor}
\end{subfigure}%
\caption{LDA plot of brain imaging data by site index. Cluster 1 consists of site 3, 6 and 9. Cluster 4 consists of site 1 and 8.}
$^{[b]}$colored by site index, $^{[c]}$colored by cluster index
\label{fig:SiteEffect}
\end{figure}

Furthermore, we visualize the distributions of DTI imaging features with or without harmonization. 
We perform the supervised dimension reduction technique Linear Discriminant Analysis (LDA) using site/cluster index as the target variable to reduce 228-dim DTI features to a lower dimensional space with only 2 dimensions. 
Figure~\ref{fig:SiteEffect} presents the result using site index as the target variable for site effect visualization, and Figure~\ref{fig:ClusterEffect} presents the result using cluster index as the target variable for cluster effect visualization.

For cluster effect visualization in Figure~\ref{fig:ClusterEffect}, we only colored data samples by cluster index. 
As shown in Figure~\ref{fig:ClusterEffect-unharm}, data without harmonization reveals a clear distinguishable cluster pattern, especially for cluster 3 and cluster 4, and samples of each cluster are centered around their own cluster centroid. 
This indicates that cluster effect does exist in DTI imaging. 
In both Figure~\ref{fig:ClusterEffect-combat} and~\ref{fig:ClusterEffect-cluster-combat}, the distribution of samples presents more like a single spherical shape, and different clusters overlap with each other after harmonization, which makes it harder to distinguish one from others compared with the unharmonized result. 
This suggests that harmonization methods effectively removed the cluster effect from raw DTI data.

For site effect visualization, we only show distributions of cluster 1 and cluster 4 for demonstration.
And we color the same site-index-based LDA visualization~\footnote{LDA visualization results will differ depending on the choice of target variable.} using different coloring strategies, for a better understanding of relations between site and cluster effect in \textit{Cluster ComBat}: the left column figures (Figure~\ref{fig:SiteEffect-unharm-sitecolor}, \ref{fig:SiteEffect-clustercombat-sitecolor}) are colored by site index, while the right column figures (Figure~\ref{fig:SiteEffect-unharm-clustercolor}, \ref{fig:SiteEffect-clustercombat-clustercolor}) are colored by cluster index. 
By comparing Figure~\ref{fig:SiteEffect-unharm-sitecolor} and \ref{fig:SiteEffect-unharm-clustercolor}, we can know that both site effect and cluster effect are evident in unharmonized data, as distinct separation is observed between sites and clusters.
By comparing Figure~\ref{fig:SiteEffect-unharm-clustercolor} and \ref{fig:SiteEffect-clustercombat-clustercolor}, we verify that our \textit{Cluster ComBat} does remove cluster effect, as cluster 1 and cluster 4 overlap with each other after harmonization. 
Then, by coloring samples in the same cluster differently based on site index, as shown in Figure~\ref{fig:SiteEffect-clustercombat-sitecolor}, we find that cluster 1 consists of site 3, 6, 9 and 12. 
Though site 6 and 12 overlap with each other, site 3, 6 and 9 are clearly separated from each other. 
Similar to cluster 4, site 1 and 8 show obvious disparity with each other. 
To conclude, our \textit{Cluster ComBat} removes differences over clusters while preserving possible site differences within the cluster, which is also verified in higher site classification accuracy than \textsc{ComBat} in Table~\ref{tab:siteClass}.


\subsection{Downstream Regression Performance}\label{congitiveRegressionTask}

For real data, we do not have the ground truth of harmonization, so our focus is on evaluating the performance of harmonization algorithms through downstream tasks. 
In these tasks, we use the 228 features of DTI brain imaging to predict the MEM, MEM SLOPES, EXF, EXF SLOPES, LAN, and LAN SLOPES variables. 
We build a simple Linear Regression model using the Scikit-Learn library \cite{scikit-learn} to train the regression task on the target target variables.
For \textsc{ComBat} and Distributed ComBat, we retrain parameters as described in Section~\ref{Simlutation}. 
We split the 18 sites into 12 training sites and 6 testing sites, then run experiments 100 times with different combinations of train and test sites.
To evaluate performance, we compute the Mean Absolute Error (MAE) of the linear regression's outputs on the testing site's data and target testing labels. In a centralized setting, we also compared our method with the Generalized Linear Squares Approach~\cite{Wang2018}, an algorithm designed to eliminate confounding effects. This approach assumes that a variable may be linearly dependent on the confounding variables, and these effects can be removed by solving a linear regression optimization problem.
Results in Table~\ref{tab:brain1} show that our proposed method performs better than \textsc{ComBat} and Generalized Linear Squares Approach in a centralized setting and Distributed ComBat in a decentralized setting for most downstream tasks.

\begin{table*}[!ht]
\small
  \caption[Table]{Performance of downstream regression task for neuroimaging dataset. $^{[a]}$ means retraining with test sites.}\label{tab:brain1}
  \centering
  \begin{threeparttable}
  \begin{tabular}{c|cccccc}
    \toprule
    Algorithm & MEM & MEM SLOPES & EXF & EXF SLOPES & LAN & LAN SLOPES \\
    \midrule
    \multicolumn{7}{l}{Centralized Setting} \\
    \midrule
    Without harmonization & 13.77±22.05 & 1.89±3.59 & 10.30±19.38& 1.58±3.19 & 10.94±17.74 & 1.45±3.01\\
    
   Generalized Linear Squares Approach ~\cite{Wang2018} & 1.07±0.30 & 0.52±0.18 & 0.93±0.22 & 0.47±0.18 & 0.95±0.26 & 0.45±0.13\\
    
    \textsc{ComBat}$^{[a]}$ & \textbf{1.00±0.18} & 0.16±0.04 &  1.03±0.18 &0.13±0.04 &  1.04±0.20 &0.13±0.03\\

    \textit{Cluster ComBat} &  1.00±0.20 &   \textbf{0.15±0.03} & \textbf{0.91±0.12} & \textbf{0.12±0.03} & \textbf{0.87±0.15} & \textbf{0.12±0.02} \\
    \midrule
    \multicolumn{7}{l}{Decentralized Setting} \\
    \midrule
    Distributed ComBat$^{[a]}$ & 0.98±0.16 & 0.15±0.03 & 1.00±0.16 & 0.13±0.03 &1.01±0.17 & 0.12±0.03\\
    Distributed \textit{Cluster ComBat} & \textbf{0.91±0.16} &  \textbf{0.14±0.03} & \textbf{0.96±0.12} & \textbf{0.12±0.02} & \textbf{0.91±0.17} &  \textbf{0.11±0.02}\\
  \bottomrule
\end{tabular}
    
    \end{threeparttable}
\end{table*}

\begin{table*}[!ht]
\small
  \caption{Effect of number of clusters $k$ on \textit{Cluster ComBat} for downstream regression task}
  \label{tab:differentK}
  \begin{tabular}{c|cccccc}
    \toprule
  \multirow{2}{*}{k} & MEM & MEM SLOPES & EXF & EXF SLOPES & LAN & LAN SLOPES \\
   & \multicolumn{6}{c}{(Centralized setting / Decentralized setting)}\\
    \midrule
     3 & 1.06±0.35 / 0.91±0.16 & 0.16±0.07 / 0.14±0.03  & 0.93±0.25 / 0.96±0.12 & 0.13±0.03 / 0.12±0.02 & 0.93±0.23 / 0.91±0.17 & 0.13±0.05 / 0.11±0.02\\
    5  & 1.00±0.20 / 0.93±0.16 & 0.15±0.03 / 0.14±0.03 & 0.91±0.12 / 0.97±0.13 & 0.12±0.03 / 0.12±0.02&  0.87±0.15 / 0.90±0.14 &  0.12±0.02 / 0.12±0.02 \\
    7 &  1.02±0.24 / 0.96±0.17 & 0.16±0.04 / 0.14±0.03 & 0.92±0.22 / 0.98±0.13 & 0.12±0.02 / 0.12±0.02 & 0.91±0.18 / 0.92±0.14 &   0.13±0.03 / 0.12±0.02\\
    9  & 1.07±0.23 / 1.01±0.19 & 0.17±0.03 / 0.15±0.03 & 0.94±0.18 / 1.02±0.14 & 0.13±0.03 / 0.13±0.02 & 0.93±0.18 / 0.97±0.17 & 0.14±0.03 / 0.13±0.02 \\
  \bottomrule
\end{tabular}
\end{table*}

\subsection{Additional Empirical Studies}
\noindent\textbf{Time complexity efficiency.} To demonstrate that our proposed \textit{Cluster ComBat} does show better time efficiency compared with \textsc{ComBat} in both centralized and decentralized settings, we provide an empirical comparison of computation time. We evaluated the average running time (in seconds) for predicting MEM regression results using the ADNI dataset in 100 experiments. As shown in Table~\ref{tab:timeefficiency}, \textit{Cluster ComBat} consistently outperforms the original \textsc{ComBat} in terms of running time, $2\times$ faster in the centralized setting and $4\times$ faster in the decentralized setting.
\begin{table}[!ht]
  \caption{Time efficiency of harmonization algorithms for MEM regression task.  $^{[a]}$ means retraining with test sites.}
  \label{tab:timeefficiency}
  \centering
  \begin{threeparttable}
  \begin{tabular}{c|c}
  
    \toprule
     Algorithm & Average Time (s) \\
     \midrule
    \multicolumn{2}{l}{Centralized Setting} \\
    
    \midrule
    \textsc{ComBat}$^{[a]}$ & 0.2427±.0.0017 \\
    \textit{Cluster ComBat} & 0.1127±.0.0001  \\
\midrule
    \multicolumn{2}{l}{Decentralized Setting} \\
    
    \midrule
    Distributed ComBat$^{[a]}$ & 	2.5051±0.0771  \\
    Distributed \textit{Cluster ComBat} & 0.6389±.0.0027 \\
    
  \bottomrule
  
\end{tabular}
    
    \end{threeparttable}
\end{table}

\noindent\textbf{Number of Clusters.}
We investigate the impact of the number of clusters ($k$) for K-means on both \textit{Cluster ComBat} and \textit{Distributed Cluster ComBat}. 
We conduct the same downstream tasks experiments as described in Section ~\ref{congitiveRegressionTask} with different numbers of clusters for the K-means algorithm, specifically $3, 5, 7,$ and $9$. Average performances are reported in Table \ref{tab:differentK}. As observed in Table \ref{tab:differentK}, variations in the number of clusters ($k$) do not significantly affect the regression performance across $100$ different random seed experiments for all six target variables. This indicates that our \textit{Cluster ComBat} methods are stable among different numbers of clusters.

\noindent\textbf{Limited Sample Size Per Sites.}
One advantage of our proposed methods is that they can still harmonize data even in limited sample sizes at each site. 
This is attributed to the fact that we have larger samples in clusters instead of individual sites. 
We investigated this by restricting the selection to a maximum number of samples at each site, such as $10, 20, 40, 60$. 
We performed a regression task over the EXF variable, and the average performance of 100 experiments is reported in Table~\ref{tab:limitedSampleSize}. We see that when the sample size is limited to $10$, \textsc{ComBat} fails to harmonize. 
However, our proposed \textit{Cluster ComBat} still achieves comparable regression performance without harmonization. 
For maximum sample sizes per site of $20, 40, 60$, our proposed method consistently outperforms the baseline \textsc{ComBat}.

\begin{table}[!ht]
\small
\setlength\tabcolsep{0.2em}

  \caption{Effect of limiting number of samples $n$ per site for harmonization methods. $^{[a]}$ means retraining with test sites.}
  \label{tab:limitedSampleSize}
  \centering
  \begin{threeparttable}
  \begin{tabular}{c|cccc}
  
    \toprule
     Algorithm & $n = 10$ & $n = 20$  & $n = 40$ &  $n= 60$  \\
    \midrule
    Without harmonization & 0.73±0.15 &33.60±79.37 & 20.45±39.30 &10.84±20.11   \\
    \textsc{ComBat}$^{[a]}$ & 2.03±0.70 & 2.71±1.09  &  1.10±0.22 &  1.06±0.18 \\
    \textit{Cluster ComBat} & 0.80±0.16 & 2.36±1.11 & 0.97±0.28 & 0.91±0.14 \\
  \bottomrule
  
\end{tabular}
    
    \end{threeparttable}
\end{table}

\noindent\textbf{Important Feature Before and After Harmonization.}
For regression tasks, we compute $p$-values for linear regression across 228 features in DTI imaging. The final $p$-values are obtained by averaging over 100 different random seeds. A feature is important if its $p$-value is less than 0.05. Table ~\ref{tab:importantfeature} displays the number of important features for the linear regression across 3 target variables MEM, EXF, and LAN. The table indicates that by using \textit{Cluster ComBat}, we achieve comparable performance with fewer significant features. This suggests that without harmonization and \textsc{ComBat}, the model may have included too many false positive features.

\begin{figure}[!t]

\begin{subfigure}{.5\linewidth}
  \centering
  \includegraphics[width=\linewidth]{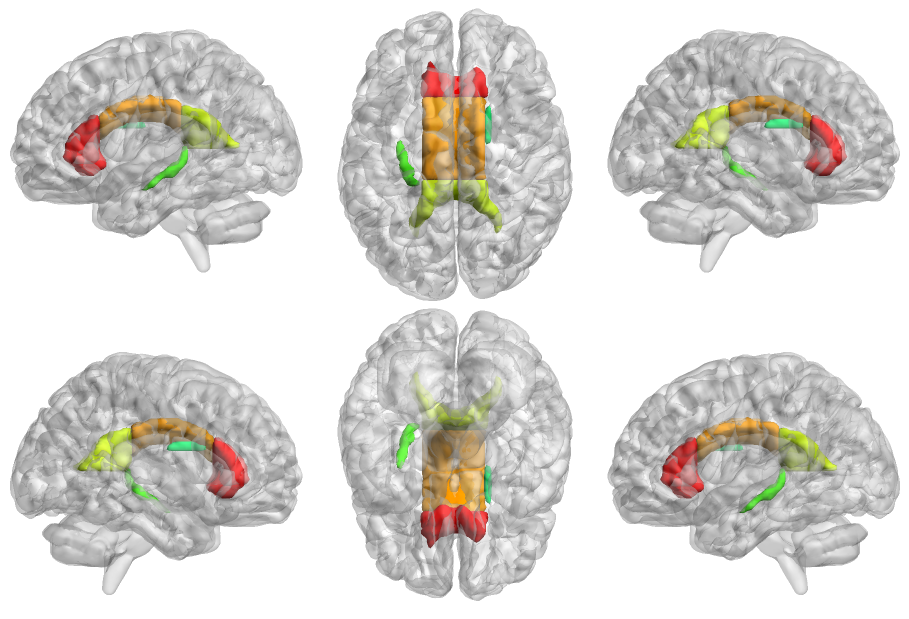}
  \caption{MEM}
\end{subfigure}%
\begin{subfigure}{.5\linewidth}
  \centering
 \includegraphics[width=\linewidth]{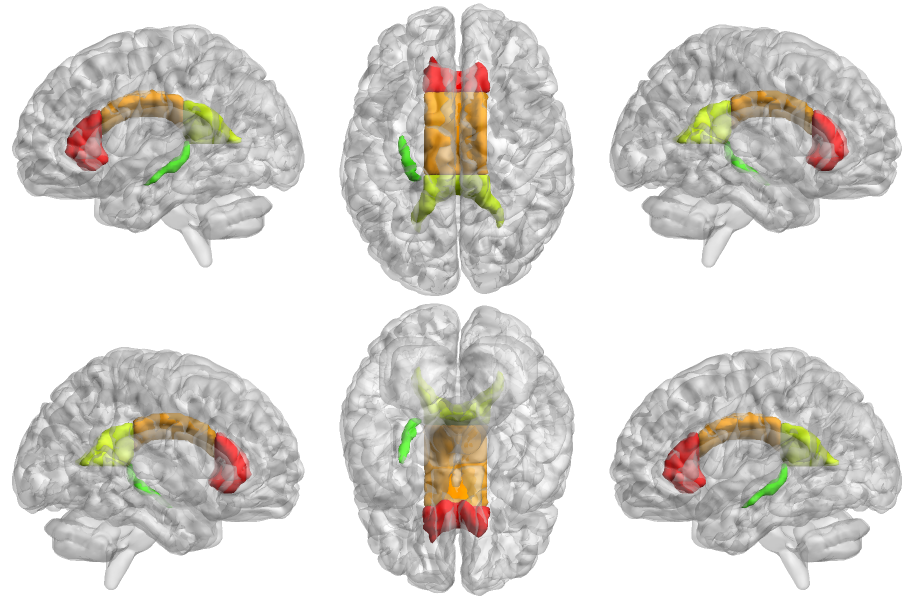}
  \caption{EXF}
\end{subfigure}%

\begin{subfigure}{.5\linewidth}
  \centering
  \includegraphics[width=\linewidth]{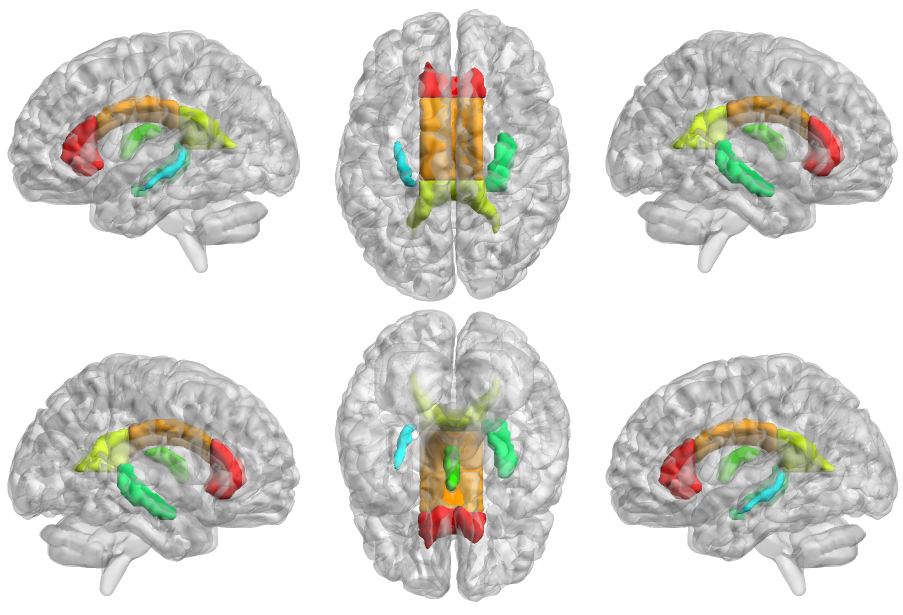}
  \caption{LAN}
\end{subfigure}%
\caption{Important Feature Visualization. The left fornix (cres)-stria terminalis and the right superior fronto-occipital fasciculus play a role in MEM. In EXF, the involvement includes the left fornix (cres)-stria terminalis and the full corpus callosum on both sides. For LAN, the engagement extends to the bilateral fornix, full corpus callosum, and bilateral fornix (cres)-stria terminalis. The color in the figure serves solely to distinguish the Regions of Interest (ROIs).}
\label{fig:ClusterImportantFeautre}
\end{figure}

In addition to $p$-values, another measure of feature importance is provided by the linear regression coefficients. The magnitude of the coefficient indicates the importance of a feature. Similar to the approach used for deriving final $p$-values, we compute the average of linear regression coefficients across experiments. Our findings highlight the significant involvement of multiple fiber tracts, such as the fornix(cres)-stria terminalis, superior fronto-occipital fasciculus, corpus callosum, and fornix, in three cognitive tasks: MEM (memory), LAN (language), and EXF (emotion), which are consistent with existing literature~\cite{zahr2009problem,frederiksen2013corpus,johnson2017endothelial,seer2022bridging,bourbon2023associations,liu2024bridging}. Figure ~\ref{fig:ClusterImportantFeautre} visualizes the details of fiber tracts for each cognitive task.

\begin{table}[!ht]
  \caption{Number of important features $(p < 0.05)$ for MEM, EXF, LAN regression task. $^{[a]}$ means retraining with test sites.}
  \label{tab:importantfeature}
  \centering
  \begin{threeparttable}
  \begin{tabular}{c|ccc}
  
    \toprule
     Algorithm & MEM  & EXF &  LAN  \\
    \midrule
    Without harmonization &  150 &  127 &  147  \\
    \textsc{ComBat}$^{[a]}$ & 156 &  135 &  150 \\
    \textit{Cluster ComBat} & 26 & 31 & 27 \\
  \bottomrule
  
\end{tabular}
    
    \end{threeparttable}
\end{table}

\section{Discussion and Conclusion}

\textsc{ComBat} has been the standard protocol for harmonization batch effects for various biomedical data analyses, and yet current \textsc{ComBat} implementations and variants cannot handle new/unseen sites, once the harmonization is done. 
In this paper, we developed a novel \textit{Cluster ComBat} and a distributed variant \textit{Distributed Cluster ComBat} to perform privacy-aware harmonization over distributed data sources and handle generalization to data in unseen sites/institutions after the harmonization is completed.   
Our proposed approach is largely aligned with existing harmonization protocols and can be easily adapted to extend harmonization to large-scale, multi-site data analyses and greatly reduce the logistic overhead of initiating distributed computing when new sites continuously join analyses. 
We believe this approach can greatly advance data-driven scientific research in multi-institutional studies, especially in the medical and biomedical domains. For example, the research activities~\cite{schijven2023large} in ENIGMA Neuroimaging Consortium~\cite{thompson2014enigma} can greatly benefit from this research when new institutions join the consortium and participate in existing studies.

We conducted extensive validation on both synthetic data real brain imaging data from ADNI in both centralized and decentralized settings. 
We demonstrated through both qualitative and quantitative studies that our methods effectively remove cluster-wise effects from brain imaging data. 
Then, our methods exhibit superior performance on downstream regression tasks compared to baseline harmonization methods in both centralized and decentralized settings, which further validates the efficacy of our harmonized data. 
We also showed that our methods can use much fewer significant features to achieve similar regression performance compared with other baselines, suggesting potential avenues for further research on selected features.

Regarding deploying our proposed method, we consider 3 implementation consideration aspects: 
 \begin{inparaenum}[1)] 
 \quad\item \textit{ML-framework agnostic:} Our algorithm doesn't involve any specific ML frameworks in the local computation part, so it is easy to implement in many systems regardless of the local ML framework. Flower~\cite{beutel2022flower} can be a candidate choice. For downstream tasks after harmonization, like regression or other ML models, the choice of local ML framework can be flexible depending on local preference. 
 \quad\item \textit{Security communication:} Designed for medical records, the deployment system needs to have communication security to prevent privacy leakage. One possible choice is to encrypt the communication between the clients and the server, for example, Secure Socket Layer (SSL)~\cite{SSL} or Transport Layer Security (TLS)~\cite{TLS}. 
 \quad\item \textit{Scalable and light-weight:} Since our algorithm’s main benefit lies in new clients joining the federated system, the system deployment should be scalable. To be more specific, when new clients join in, there should be minimum system configuration modification on the server as well as for old clients. Also, the implementation of our algorithm needs to be lightweight, and the FL system with our algorithm should require limited system consumption.  And the design of FedLab~\cite{Fedlab} can be a reference to meet these requirements. \end{inparaenum} As a future work, we will deploy our proposed \textit{Cluster ComBat} harmonization in the ENIGMA Consortium toolbox to further validate existing studies.
\section{Acknowledgement}
This material is based in part upon work supported by the National Science Foundation under
Grant IIS-2212174, IIS-1749940, IIS 2319450, IIS 2045848, Office of Naval Research N00014-24-1-2168, and National
Institute on Aging (NIA) RF1AG072449, U01AG068057, National Institute of Mental Health RF1MH125928.

\clearpage
\bibliographystyle{ACM-Reference-Format}
\balance
\bibliography{sources}


\begin{thebibliography}{57}


\ifx \showCODEN    \undefined \def \showCODEN     #1{\unskip}     \fi
\ifx \showDOI      \undefined \def \showDOI       #1{#1}\fi
\ifx \showISBNx    \undefined \def \showISBNx     #1{\unskip}     \fi
\ifx \showISBNxiii \undefined \def \showISBNxiii  #1{\unskip}     \fi
\ifx \showISSN     \undefined \def \showISSN      #1{\unskip}     \fi
\ifx \showLCCN     \undefined \def \showLCCN      #1{\unskip}     \fi
\ifx \shownote     \undefined \def \shownote      #1{#1}          \fi
\ifx \showarticletitle \undefined \def \showarticletitle #1{#1}   \fi
\ifx \showURL      \undefined \def \showURL       {\relax}        \fi
\providecommand\bibfield[2]{#2}
\providecommand\bibinfo[2]{#2}
\providecommand\natexlab[1]{#1}
\providecommand\showeprint[2][]{arXiv:#2}

\bibitem[Abdusalomov et~al\mbox{.}(2023)]%
        {Abdusalomov2023}
\bibfield{author}{\bibinfo{person}{Akmalbek~Bobomirzaevich Abdusalomov}, \bibinfo{person}{Mukhriddin Mukhiddinov}, {and} \bibinfo{person}{Taeg~Keun Whangbo}.} \bibinfo{year}{2023}\natexlab{}.
\newblock \showarticletitle{Brain Tumor Detection Based on Deep Learning Approaches and Magnetic Resonance Imaging}.
\newblock \bibinfo{journal}{\emph{Cancers}} \bibinfo{volume}{15}, \bibinfo{number}{16} (\bibinfo{date}{Aug.} \bibinfo{year}{2023}), \bibinfo{pages}{4172}.
\newblock
\showISSN{2072-6694}
\urldef\tempurl%
\url{https://doi.org/10.3390/cancers15164172}
\showDOI{\tempurl}


\bibitem[Bayer et~al\mbox{.}(2022)]%
        {Bayer2022}
\bibfield{author}{\bibinfo{person}{Johanna M.~M. Bayer}, \bibinfo{person}{Paul~M. Thompson}, \bibinfo{person}{Christopher R.~K. Ching}, \bibinfo{person}{Mengting Liu}, \bibinfo{person}{Andrew Chen}, \bibinfo{person}{Alana~C. Panzenhagen}, \bibinfo{person}{Neda Jahanshad}, \bibinfo{person}{Andre Marquand}, \bibinfo{person}{Lianne Schmaal}, {and} \bibinfo{person}{Philipp~G. S\"{a}mann}.} \bibinfo{year}{2022}\natexlab{}.
\newblock \showarticletitle{Site effects how-to and when: An overview of retrospective techniques to accommodate site effects in multi-site neuroimaging analyses}.
\newblock \bibinfo{journal}{\emph{Frontiers in Neurology}}  \bibinfo{volume}{13} (\bibinfo{date}{Oct.} \bibinfo{year}{2022}).
\newblock
\showISSN{1664-2295}
\urldef\tempurl%
\url{https://doi.org/10.3389/fneur.2022.923988}
\showDOI{\tempurl}


\bibitem[Beer et~al\mbox{.}(2020)]%
        {Beer2020}
\bibfield{author}{\bibinfo{person}{Joanne~C. Beer}, \bibinfo{person}{Nicholas~J. Tustison}, \bibinfo{person}{Philip~A. Cook}, \bibinfo{person}{Christos Davatzikos}, \bibinfo{person}{Yvette~I. Sheline}, \bibinfo{person}{Russell~T. Shinohara}, {and} \bibinfo{person}{Kristin~A. Linn}.} \bibinfo{year}{2020}\natexlab{}.
\newblock \showarticletitle{Longitudinal ComBat: A method for harmonizing longitudinal multi-scanner imaging data}.
\newblock \bibinfo{journal}{\emph{NeuroImage}}  \bibinfo{volume}{220} (\bibinfo{date}{Oct.} \bibinfo{year}{2020}), \bibinfo{pages}{117129}.
\newblock
\showISSN{1053-8119}
\urldef\tempurl%
\url{https://doi.org/10.1016/j.neuroimage.2020.117129}
\showDOI{\tempurl}


\bibitem[Beutel et~al\mbox{.}(2022)]%
        {beutel2022flower}
\bibfield{author}{\bibinfo{person}{Daniel~J. Beutel}, \bibinfo{person}{Taner Topal}, \bibinfo{person}{Akhil Mathur}, \bibinfo{person}{Xinchi Qiu}, \bibinfo{person}{Javier Fernandez-Marques}, \bibinfo{person}{Yan Gao}, \bibinfo{person}{Lorenzo Sani}, \bibinfo{person}{Kwing~Hei Li}, \bibinfo{person}{Titouan Parcollet}, \bibinfo{person}{Pedro Porto~Buarque de Gusmão}, {and} \bibinfo{person}{Nicholas~D. Lane}.} \bibinfo{year}{2022}\natexlab{}.
\newblock \bibinfo{title}{Flower: A Friendly Federated Learning Research Framework}.
\newblock
\newblock
\showeprint[arxiv]{2007.14390}~[cs.LG]


\bibitem[Bourbon-Teles et~al\mbox{.}(2023)]%
        {bourbon2023associations}
\bibfield{author}{\bibinfo{person}{Jos{\'e} Bourbon-Teles}, \bibinfo{person}{L{\'\i}lia Jorge}, \bibinfo{person}{N{\'a}dia Can{\'a}rio}, \bibinfo{person}{Ricardo Martins}, \bibinfo{person}{Isabel Santana}, {and} \bibinfo{person}{Miguel Castelo-Branco}.} \bibinfo{year}{2023}\natexlab{}.
\newblock \showarticletitle{Associations between cortical $\beta$-amyloid burden, fornix microstructure and cognitive processing of faces, places, bodies and other visual objects in early Alzheimer's disease}.
\newblock \bibinfo{journal}{\emph{Hippocampus}} \bibinfo{volume}{33}, \bibinfo{number}{2} (\bibinfo{year}{2023}), \bibinfo{pages}{112--124}.
\newblock


\bibitem[Chen et~al\mbox{.}(2022)]%
        {Chen2022}
\bibfield{author}{\bibinfo{person}{Andrew~A. Chen}, \bibinfo{person}{Chongliang Luo}, \bibinfo{person}{Yong Chen}, \bibinfo{person}{Russell~T. Shinohara}, {and} \bibinfo{person}{Haochang Shou}.} \bibinfo{year}{2022}\natexlab{}.
\newblock \showarticletitle{Privacy-preserving harmonization via distributed ComBat}.
\newblock \bibinfo{journal}{\emph{NeuroImage}}  \bibinfo{volume}{248} (\bibinfo{date}{March} \bibinfo{year}{2022}), \bibinfo{pages}{118822}.
\newblock
\showISSN{1053-8119}
\urldef\tempurl%
\url{https://doi.org/10.1016/j.neuroimage.2021.118822}
\showDOI{\tempurl}


\bibitem[Crane et~al\mbox{.}(2012)]%
        {crane2012development}
\bibfield{author}{\bibinfo{person}{Paul~K Crane}, \bibinfo{person}{Adam Carle}, \bibinfo{person}{Laura~E Gibbons}, \bibinfo{person}{Philip Insel}, \bibinfo{person}{R~Scott Mackin}, \bibinfo{person}{Alden Gross}, \bibinfo{person}{Richard~N Jones}, \bibinfo{person}{Shubhabrata Mukherjee}, \bibinfo{person}{S~McKay Curtis}, \bibinfo{person}{Danielle Harvey}, {et~al\mbox{.}}} \bibinfo{year}{2012}\natexlab{}.
\newblock \showarticletitle{Development and assessment of a composite score for memory in the Alzheimer’s Disease Neuroimaging Initiative (ADNI)}.
\newblock \bibinfo{journal}{\emph{Brain imaging and behavior}}  \bibinfo{volume}{6} (\bibinfo{year}{2012}), \bibinfo{pages}{502--516}.
\newblock


\bibitem[Dipro et~al\mbox{.}(2022)]%
        {dipro2022federated}
\bibfield{author}{\bibinfo{person}{Sumit~Howlader Dipro}, \bibinfo{person}{Mynul Islam}, \bibinfo{person}{Abdullah Al~Nahian}, \bibinfo{person}{Moonami~Sharmita Azad}, \bibinfo{person}{Amitabha Chakrabarty}, {and} \bibinfo{person}{Tanzim Reza}.} \bibinfo{year}{2022}\natexlab{}.
\newblock \showarticletitle{A Federated Learning Based Privacy Preserving Approach for Detecting Parkinson’s Disease Using Deep Learning}. In \bibinfo{booktitle}{\emph{2022 25th International Conference on Computer and Information Technology (ICCIT)}}. IEEE, \bibinfo{pages}{139--144}.
\newblock


\bibitem[Duan et~al\mbox{.}(2019)]%
        {Duan2019}
\bibfield{author}{\bibinfo{person}{Rui Duan}, \bibinfo{person}{Mary~Regina Boland}, \bibinfo{person}{Zixuan Liu}, \bibinfo{person}{Yue Liu}, \bibinfo{person}{Howard~H Chang}, \bibinfo{person}{Hua Xu}, \bibinfo{person}{Haitao Chu}, \bibinfo{person}{Christopher~H Schmid}, \bibinfo{person}{Christopher~B Forrest}, \bibinfo{person}{John~H Holmes}, \bibinfo{person}{Martijn~J Schuemie}, \bibinfo{person}{Jesse~A Berlin}, \bibinfo{person}{Jason~H Moore}, {and} \bibinfo{person}{Yong Chen}.} \bibinfo{year}{2019}\natexlab{}.
\newblock \showarticletitle{Learning from electronic health records across multiple sites: A communication-efficient and privacy-preserving distributed algorithm}.
\newblock \bibinfo{journal}{\emph{Journal of the American Medical Informatics Association}} \bibinfo{volume}{27}, \bibinfo{number}{3} (\bibinfo{date}{Dec.} \bibinfo{year}{2019}), \bibinfo{pages}{376–385}.
\newblock
\showISSN{1527-974X}
\urldef\tempurl%
\url{https://doi.org/10.1093/jamia/ocz199}
\showDOI{\tempurl}


\bibitem[Frederiksen(2013)]%
        {frederiksen2013corpus}
\bibfield{author}{\bibinfo{person}{Kristian~Steen Frederiksen}.} \bibinfo{year}{2013}\natexlab{}.
\newblock \showarticletitle{Corpus callosum in aging and dementia}.
\newblock \bibinfo{journal}{\emph{Dan Med J}} \bibinfo{volume}{60}, \bibinfo{number}{10} (\bibinfo{year}{2013}), \bibinfo{pages}{B4721}.
\newblock


\bibitem[Fukasawa et~al\mbox{.}(2004)]%
        {Fukasawa2004}
\bibfield{author}{\bibinfo{person}{Takuya Fukasawa}, \bibinfo{person}{Jiahong Wang}, \bibinfo{person}{Toyoo Takata}, {and} \bibinfo{person}{Masatoshi Miyazaki}.} \bibinfo{year}{2004}\natexlab{}.
\newblock \bibinfo{booktitle}{\emph{An Effective Distributed Privacy-Preserving Data Mining Algorithm}}.
\newblock \bibinfo{publisher}{Springer Berlin Heidelberg}, \bibinfo{pages}{320–325}.
\newblock
\showISBNx{9783540286516}
\showISSN{1611-3349}
\urldef\tempurl%
\url{https://doi.org/10.1007/978-3-540-28651-6_47}
\showDOI{\tempurl}


\bibitem[Garcia Santa~Cruz et~al\mbox{.}(2023)]%
        {GarciaSantaCruz2023}
\bibfield{author}{\bibinfo{person}{Beatriz Garcia Santa~Cruz}, \bibinfo{person}{Andreas Husch}, {and} \bibinfo{person}{Frank Hertel}.} \bibinfo{year}{2023}\natexlab{}.
\newblock \showarticletitle{Machine learning models for diagnosis and prognosis of Parkinson’s disease using brain imaging: general overview, main challenges, and future directions}.
\newblock \bibinfo{journal}{\emph{Frontiers in Aging Neuroscience}}  \bibinfo{volume}{15} (\bibinfo{date}{July} \bibinfo{year}{2023}).
\newblock
\showISSN{1663-4365}
\urldef\tempurl%
\url{https://doi.org/10.3389/fnagi.2023.1216163}
\showDOI{\tempurl}


\bibitem[Gibbons et~al\mbox{.}(2012)]%
        {gibbons2012composite}
\bibfield{author}{\bibinfo{person}{Laura~E Gibbons}, \bibinfo{person}{Adam~C Carle}, \bibinfo{person}{R~Scott Mackin}, \bibinfo{person}{S Mukherjee}, \bibinfo{person}{P Insel}, \bibinfo{person}{SM Curtis}, \bibinfo{person}{A Gross}, \bibinfo{person}{RN Jones}, \bibinfo{person}{D Mungas}, \bibinfo{person}{M Weiner}, {et~al\mbox{.}}} \bibinfo{year}{2012}\natexlab{}.
\newblock \showarticletitle{Composite measures of executive function and memory: ADNI\_EF and ADNI\_Mem}.
\newblock \bibinfo{journal}{\emph{Alzheimer’s Dis Neuroimaging Initiat}} (\bibinfo{year}{2012}).
\newblock


\bibitem[Haque et~al\mbox{.}(2023)]%
        {Haque2023}
\bibfield{author}{\bibinfo{person}{Rakib~Ul Haque}, \bibinfo{person}{A.S.M.~Touhidul Hasan}, \bibinfo{person}{Apubra Daria}, \bibinfo{person}{Abdur Rasool}, \bibinfo{person}{Hui Chen}, \bibinfo{person}{Qingshan Jiang}, {and} \bibinfo{person}{Yuqing Zhang}.} \bibinfo{year}{2023}\natexlab{}.
\newblock \showarticletitle{A novel secure and distributed architecture for privacy-preserving healthcare system}.
\newblock \bibinfo{journal}{\emph{Journal of Network and Computer Applications}}  \bibinfo{volume}{217} (\bibinfo{date}{Aug.} \bibinfo{year}{2023}), \bibinfo{pages}{103696}.
\newblock
\showISSN{1084-8045}
\urldef\tempurl%
\url{https://doi.org/10.1016/j.jnca.2023.103696}
\showDOI{\tempurl}


\bibitem[Hidayat et~al\mbox{.}(2023)]%
        {SSL}
\bibfield{author}{\bibinfo{person}{Muhammad Hidayat}, \bibinfo{person}{Yugo Nakamura}, {and} \bibinfo{person}{Yutaka Arakawa}.} \bibinfo{year}{2023}\natexlab{}.
\newblock \showarticletitle{Privacy-Preserving Federated Learning With Resource Adaptive Compression for Edge Devices}.
\newblock \bibinfo{journal}{\emph{IEEE Internet of Things Journal}}  \bibinfo{volume}{PP} (\bibinfo{date}{01} \bibinfo{year}{2023}), \bibinfo{pages}{1--1}.
\newblock
\urldef\tempurl%
\url{https://doi.org/10.1109/JIOT.2023.3347552}
\showDOI{\tempurl}


\bibitem[Hohman et~al\mbox{.}(2017)]%
        {hohman2017evaluating}
\bibfield{author}{\bibinfo{person}{Timothy~J Hohman}, \bibinfo{person}{Doug Tommet}, \bibinfo{person}{Shawn Marks}, \bibinfo{person}{Joey Contreras}, \bibinfo{person}{Rich Jones}, \bibinfo{person}{Dan Mungas}, \bibinfo{person}{Alzheimer's~Neuroimaging Initiative}, {et~al\mbox{.}}} \bibinfo{year}{2017}\natexlab{}.
\newblock \showarticletitle{Evaluating Alzheimer's disease biomarkers as mediators of age-related cognitive decline}.
\newblock \bibinfo{journal}{\emph{Neurobiology of aging}}  \bibinfo{volume}{58} (\bibinfo{year}{2017}), \bibinfo{pages}{120--128}.
\newblock


\bibitem[Hong et~al\mbox{.}(2021)]%
        {hong2021learning}
\bibfield{author}{\bibinfo{person}{Junyuan Hong}, \bibinfo{person}{Haotao Wang}, \bibinfo{person}{Zhangyang Wang}, {and} \bibinfo{person}{Jiayu Zhou}.} \bibinfo{year}{2021}\natexlab{}.
\newblock \showarticletitle{Learning model-based privacy protection under budget constraints}. In \bibinfo{booktitle}{\emph{Proceedings of the AAAI Conference on Artificial Intelligence}}, Vol.~\bibinfo{volume}{35}. \bibinfo{pages}{7702--7710}.
\newblock


\bibitem[Hong et~al\mbox{.}(2022)]%
        {hong2022dynamic}
\bibfield{author}{\bibinfo{person}{Junyuan Hong}, \bibinfo{person}{Zhangyang Wang}, {and} \bibinfo{person}{Jiayu Zhou}.} \bibinfo{year}{2022}\natexlab{}.
\newblock \showarticletitle{Dynamic privacy budget allocation improves data efficiency of differentially private gradient descent}. In \bibinfo{booktitle}{\emph{Proceedings of the 2022 ACM Conference on Fairness, Accountability, and Transparency}}. \bibinfo{pages}{11--35}.
\newblock


\bibitem[Jalali and Chen(2024)]%
        {TLS}
\bibfield{author}{\bibinfo{person}{Nasir~Ahmad Jalali} {and} \bibinfo{person}{Hongsong Chen}.} \bibinfo{year}{2024}\natexlab{}.
\newblock \showarticletitle{Federated Learning Security and Privacy-Preserving Algorithm and Experiments Research Under Internet of Things Critical Infrastructure}.
\newblock \bibinfo{journal}{\emph{Tsinghua Science and Technology}} \bibinfo{volume}{29}, \bibinfo{number}{2} (\bibinfo{year}{2024}), \bibinfo{pages}{400--414}.
\newblock
\urldef\tempurl%
\url{https://doi.org/10.26599/TST.2023.9010007}
\showDOI{\tempurl}


\bibitem[Johnson et~al\mbox{.}(2017)]%
        {johnson2017endothelial}
\bibfield{author}{\bibinfo{person}{Nathan~F Johnson}, \bibinfo{person}{Brian~T Gold}, \bibinfo{person}{Christopher~A Brown}, \bibinfo{person}{Emily~F Anggelis}, \bibinfo{person}{Alison~L Bailey}, \bibinfo{person}{Jody~L Clasey}, {and} \bibinfo{person}{David~K Powell}.} \bibinfo{year}{2017}\natexlab{}.
\newblock \showarticletitle{Endothelial function is associated with white matter microstructure and executive function in older adults}.
\newblock \bibinfo{journal}{\emph{Frontiers in Aging Neuroscience}}  \bibinfo{volume}{9} (\bibinfo{year}{2017}), \bibinfo{pages}{255}.
\newblock


\bibitem[Johnson et~al\mbox{.}(2006)]%
        {Johnson2006}
\bibfield{author}{\bibinfo{person}{W.~Evan Johnson}, \bibinfo{person}{Cheng Li}, {and} \bibinfo{person}{Ariel Rabinovic}.} \bibinfo{year}{2006}\natexlab{}.
\newblock \showarticletitle{Adjusting batch effects in microarray expression data using empirical Bayes methods}.
\newblock \bibinfo{journal}{\emph{Biostatistics}} \bibinfo{volume}{8}, \bibinfo{number}{1} (\bibinfo{date}{April} \bibinfo{year}{2006}), \bibinfo{pages}{118–127}.
\newblock
\showISSN{1465-4644}
\urldef\tempurl%
\url{https://doi.org/10.1093/biostatistics/kxj037}
\showDOI{\tempurl}


\bibitem[Li et~al\mbox{.}(2021)]%
        {Li2021}
\bibfield{author}{\bibinfo{person}{Qiongxiu Li}, \bibinfo{person}{Jaron~Skovsted Gundersen}, \bibinfo{person}{Richard Heusdens}, {and} \bibinfo{person}{Mads~Grasboll Christensen}.} \bibinfo{year}{2021}\natexlab{}.
\newblock \showarticletitle{Privacy-Preserving Distributed Processing: Metrics, Bounds and Algorithms}.
\newblock \bibinfo{journal}{\emph{IEEE Transactions on Information Forensics and Security}}  \bibinfo{volume}{16} (\bibinfo{year}{2021}), \bibinfo{pages}{2090–2103}.
\newblock
\showISSN{1556-6021}
\urldef\tempurl%
\url{https://doi.org/10.1109/tifs.2021.3050064}
\showDOI{\tempurl}


\bibitem[Liu et~al\mbox{.}(2024)]%
        {liu2024bridging}
\bibfield{author}{\bibinfo{person}{Shan-Wen Liu}, \bibinfo{person}{Xiao-Ting Ma}, \bibinfo{person}{Shuai Yu}, \bibinfo{person}{Xiao-Fen Weng}, \bibinfo{person}{Meng Li}, \bibinfo{person}{Jiangtao Zhu}, \bibinfo{person}{Chun-Feng Liu}, {and} \bibinfo{person}{Hua Hu}.} \bibinfo{year}{2024}\natexlab{}.
\newblock \showarticletitle{Bridging Reduced Grip Strength and Altered Executive Function: Specific Brain White Matter Structural Changes in Patients with Alzheimer’s Disease}.
\newblock \bibinfo{journal}{\emph{Clinical Interventions in Aging}} (\bibinfo{year}{2024}), \bibinfo{pages}{93--107}.
\newblock


\bibitem[Maćkiewicz and Ratajczak(1993)]%
        {Makiewicz1993}
\bibfield{author}{\bibinfo{person}{Andrzej Maćkiewicz} {and} \bibinfo{person}{Waldemar Ratajczak}.} \bibinfo{year}{1993}\natexlab{}.
\newblock \showarticletitle{Principal components analysis (PCA)}.
\newblock \bibinfo{journal}{\emph{Computers\&amp; Geosciences}} \bibinfo{volume}{19}, \bibinfo{number}{3} (\bibinfo{date}{March} \bibinfo{year}{1993}), \bibinfo{pages}{303–342}.
\newblock
\showISSN{0098-3004}
\urldef\tempurl%
\url{https://doi.org/10.1016/0098-3004(93)90090-r}
\showDOI{\tempurl}


\bibitem[McMahan et~al\mbox{.}(2016)]%
        {FedAVG}
\bibfield{author}{\bibinfo{person}{H.~Brendan McMahan}, \bibinfo{person}{Eider Moore}, \bibinfo{person}{Daniel Ramage}, \bibinfo{person}{Seth Hampson}, {and} \bibinfo{person}{Blaise Ag\"{u}era~y Arcas}.} \bibinfo{year}{2016}\natexlab{}.
\newblock \showarticletitle{Communication-Efficient Learning of Deep Networks from Decentralized Data}.
\newblock  (\bibinfo{year}{2016}).
\newblock
\urldef\tempurl%
\url{https://doi.org/10.48550/ARXIV.1602.05629}
\showDOI{\tempurl}


\bibitem[Monsour(2022)]%
        {Monsour2022}
\bibfield{author}{\bibinfo{person}{Robert Monsour}.} \bibinfo{year}{2022}\natexlab{}.
\newblock \showarticletitle{Neuroimaging in the Era of Artificial Intelligence: Current Applications}.
\newblock \bibinfo{journal}{\emph{Federal Practitioner}} \bibinfo{number}{39 (Suppl 1)} (\bibinfo{date}{April} \bibinfo{year}{2022}).
\newblock
\urldef\tempurl%
\url{https://doi.org/10.12788/fp.0231}
\showDOI{\tempurl}


\bibitem[Nir et~al\mbox{.}(2013)]%
        {nir2013effectiveness}
\bibfield{author}{\bibinfo{person}{Talia~M Nir}, \bibinfo{person}{Neda Jahanshad}, \bibinfo{person}{Julio~E Villalon-Reina}, \bibinfo{person}{Arthur~W Toga}, \bibinfo{person}{Clifford~R Jack}, \bibinfo{person}{Michael~W Weiner}, \bibinfo{person}{Paul~M Thompson}, \bibinfo{person}{Alzheimer's Disease Neuroimaging~Initiative (ADNI}, {et~al\mbox{.}}} \bibinfo{year}{2013}\natexlab{}.
\newblock \showarticletitle{Effectiveness of regional DTI measures in distinguishing Alzheimer's disease, MCI, and normal aging}.
\newblock \bibinfo{journal}{\emph{NeuroImage: clinical}}  \bibinfo{volume}{3} (\bibinfo{year}{2013}), \bibinfo{pages}{180--195}.
\newblock


\bibitem[Orlhac et~al\mbox{.}(2021)]%
        {Orlhac2021}
\bibfield{author}{\bibinfo{person}{Fanny Orlhac}, \bibinfo{person}{Jakoba~J. Eertink}, \bibinfo{person}{Anne-Ségolène Cottereau}, \bibinfo{person}{Josée~M. Zijlstra}, \bibinfo{person}{Catherine Thieblemont}, \bibinfo{person}{Michel Meignan}, \bibinfo{person}{Ronald Boellaard}, {and} \bibinfo{person}{Irène Buvat}.} \bibinfo{year}{2021}\natexlab{}.
\newblock \showarticletitle{A Guide to ComBat Harmonization of Imaging Biomarkers in Multicenter Studies}.
\newblock \bibinfo{journal}{\emph{Journal of Nuclear Medicine}} \bibinfo{volume}{63}, \bibinfo{number}{2} (\bibinfo{date}{Sept.} \bibinfo{year}{2021}), \bibinfo{pages}{172–179}.
\newblock
\showISSN{2159-662X}
\urldef\tempurl%
\url{https://doi.org/10.2967/jnumed.121.262464}
\showDOI{\tempurl}


\bibitem[Pedregosa et~al\mbox{.}(2011)]%
        {scikit-learn}
\bibfield{author}{\bibinfo{person}{F. Pedregosa}, \bibinfo{person}{G. Varoquaux}, \bibinfo{person}{A. Gramfort}, \bibinfo{person}{V. Michel}, \bibinfo{person}{B. Thirion}, \bibinfo{person}{O. Grisel}, \bibinfo{person}{M. Blondel}, \bibinfo{person}{P. Prettenhofer}, \bibinfo{person}{R. Weiss}, \bibinfo{person}{V. Dubourg}, \bibinfo{person}{J. Vanderplas}, \bibinfo{person}{A. Passos}, \bibinfo{person}{D. Cournapeau}, \bibinfo{person}{M. Brucher}, \bibinfo{person}{M. Perrot}, {and} \bibinfo{person}{E. Duchesnay}.} \bibinfo{year}{2011}\natexlab{}.
\newblock \showarticletitle{Scikit-learn: Machine Learning in {P}ython}.
\newblock \bibinfo{journal}{\emph{Journal of Machine Learning Research}}  \bibinfo{volume}{12} (\bibinfo{year}{2011}), \bibinfo{pages}{2825--2830}.
\newblock


\bibitem[Pfaehler et~al\mbox{.}(2019)]%
        {Pfaehler2019}
\bibfield{author}{\bibinfo{person}{Elisabeth Pfaehler}, \bibinfo{person}{Joyce van Sluis}, \bibinfo{person}{Bram~B.J. Merema}, \bibinfo{person}{Peter van Ooijen}, \bibinfo{person}{Ralph~C.M. Berendsen}, \bibinfo{person}{Floris~H.P. van Velden}, {and} \bibinfo{person}{Ronald Boellaard}.} \bibinfo{year}{2019}\natexlab{}.
\newblock \showarticletitle{Experimental Multicenter and Multivendor Evaluation of the Performance of PET Radiomic Features Using 3-Dimensionally Printed Phantom Inserts}.
\newblock \bibinfo{journal}{\emph{Journal of Nuclear Medicine}} \bibinfo{volume}{61}, \bibinfo{number}{3} (\bibinfo{date}{Aug.} \bibinfo{year}{2019}), \bibinfo{pages}{469–476}.
\newblock
\showISSN{2159-662X}
\urldef\tempurl%
\url{https://doi.org/10.2967/jnumed.119.229724}
\showDOI{\tempurl}


\bibitem[Pomponio et~al\mbox{.}(2020)]%
        {Pomponio2020}
\bibfield{author}{\bibinfo{person}{Raymond Pomponio}, \bibinfo{person}{Guray Erus}, \bibinfo{person}{Mohamad Habes}, \bibinfo{person}{Jimit Doshi}, \bibinfo{person}{Dhivya Srinivasan}, \bibinfo{person}{Elizabeth Mamourian}, \bibinfo{person}{Vishnu Bashyam}, \bibinfo{person}{Ilya~M. Nasrallah}, \bibinfo{person}{Theodore~D. Satterthwaite}, \bibinfo{person}{Yong Fan}, \bibinfo{person}{Lenore~J. Launer}, \bibinfo{person}{Colin~L. Masters}, \bibinfo{person}{Paul Maruff}, \bibinfo{person}{Chuanjun Zhuo}, \bibinfo{person}{Henry V\"{o}lzke}, \bibinfo{person}{Sterling~C. Johnson}, \bibinfo{person}{Jurgen Fripp}, \bibinfo{person}{Nikolaos Koutsouleris}, \bibinfo{person}{Daniel~H. Wolf}, \bibinfo{person}{Raquel Gur}, \bibinfo{person}{Ruben Gur}, \bibinfo{person}{John Morris}, \bibinfo{person}{Marilyn~S. Albert}, \bibinfo{person}{Hans~J. Grabe}, \bibinfo{person}{Susan~M. Resnick}, \bibinfo{person}{R.~Nick Bryan}, \bibinfo{person}{David~A. Wolk}, \bibinfo{person}{Russell~T. Shinohara}, \bibinfo{person}{Haochang Shou},
  {and} \bibinfo{person}{Christos Davatzikos}.} \bibinfo{year}{2020}\natexlab{}.
\newblock \showarticletitle{Harmonization of large MRI datasets for the analysis of brain imaging patterns throughout the lifespan}.
\newblock \bibinfo{journal}{\emph{NeuroImage}}  \bibinfo{volume}{208} (\bibinfo{date}{March} \bibinfo{year}{2020}), \bibinfo{pages}{116450}.
\newblock
\showISSN{1053-8119}
\urldef\tempurl%
\url{https://doi.org/10.1016/j.neuroimage.2019.116450}
\showDOI{\tempurl}


\bibitem[Reynolds et~al\mbox{.}(2023)]%
        {Reynolds2023}
\bibfield{author}{\bibinfo{person}{Maxwell Reynolds}, \bibinfo{person}{Tigmanshu Chaudhary}, \bibinfo{person}{Mahbaneh Eshaghzadeh~Torbati}, \bibinfo{person}{Dana~L. Tudorascu}, {and} \bibinfo{person}{Kayhan Batmanghelich}.} \bibinfo{year}{2023}\natexlab{}.
\newblock \showarticletitle{ComBat Harmonization: Empirical Bayes versus fully Bayes approaches}.
\newblock \bibinfo{journal}{\emph{NeuroImage: Clinical}}  \bibinfo{volume}{39} (\bibinfo{year}{2023}), \bibinfo{pages}{103472}.
\newblock
\showISSN{2213-1582}
\urldef\tempurl%
\url{https://doi.org/10.1016/j.nicl.2023.103472}
\showDOI{\tempurl}


\bibitem[Sandhu et~al\mbox{.}(2023)]%
        {Sandhu}
\bibfield{author}{\bibinfo{person}{Sukhveer~Singh Sandhu}, \bibinfo{person}{Hamed~Taheri Gorji}, \bibinfo{person}{Pantea Tavakolian}, \bibinfo{person}{Kouhyar Tavakolian}, {and} \bibinfo{person}{Alireza Akhbardeh}.} \bibinfo{year}{2023}\natexlab{}.
\newblock \showarticletitle{Medical Imaging Applications of Federated Learning}.
\newblock \bibinfo{journal}{\emph{Diagnostics}} \bibinfo{volume}{13}, \bibinfo{number}{19} (\bibinfo{year}{2023}).
\newblock
\showISSN{2075-4418}
\urldef\tempurl%
\url{https://doi.org/10.3390/diagnostics13193140}
\showDOI{\tempurl}


\bibitem[Schijven et~al\mbox{.}(2023)]%
        {schijven2023large}
\bibfield{author}{\bibinfo{person}{Dick Schijven}, \bibinfo{person}{Merel~C Postema}, \bibinfo{person}{Masaki Fukunaga}, \bibinfo{person}{Junya Matsumoto}, \bibinfo{person}{Kenichiro Miura}, \bibinfo{person}{Sonja~MC de Zwarte}, \bibinfo{person}{Neeltje~EM Van~Haren}, \bibinfo{person}{Wiepke Cahn}, \bibinfo{person}{Hilleke~E Hulshoff~Pol}, \bibinfo{person}{Ren{\'e}~S Kahn}, {et~al\mbox{.}}} \bibinfo{year}{2023}\natexlab{}.
\newblock \showarticletitle{Large-scale analysis of structural brain asymmetries in schizophrenia via the ENIGMA consortium}.
\newblock \bibinfo{journal}{\emph{Proceedings of the National Academy of Sciences}} \bibinfo{volume}{120}, \bibinfo{number}{14} (\bibinfo{year}{2023}), \bibinfo{pages}{e2213880120}.
\newblock


\bibitem[Seer et~al\mbox{.}(2022)]%
        {seer2022bridging}
\bibfield{author}{\bibinfo{person}{Caroline Seer}, \bibinfo{person}{Hamed~Zivari Adab}, \bibinfo{person}{Justina Sidlauskaite}, \bibinfo{person}{Thijs Dhollander}, \bibinfo{person}{Sima Chalavi}, \bibinfo{person}{Jolien Gooijers}, \bibinfo{person}{Stefan Sunaert}, {and} \bibinfo{person}{Stephan~P Swinnen}.} \bibinfo{year}{2022}\natexlab{}.
\newblock \showarticletitle{Bridging cognition and action: executive functioning mediates the relationship between white matter fiber density and complex motor abilities in older adults}.
\newblock \bibinfo{journal}{\emph{Aging (Albany NY)}} \bibinfo{volume}{14}, \bibinfo{number}{18} (\bibinfo{year}{2022}), \bibinfo{pages}{7263}.
\newblock


\bibitem[Silva et~al\mbox{.}(2023)]%
        {Silva2023}
\bibfield{author}{\bibinfo{person}{Santiago Silva}, \bibinfo{person}{Neil Oxtoby}, \bibinfo{person}{Andre Altmann}, {and} \bibinfo{person}{Marco Lorenzi}.} \bibinfo{year}{2023}\natexlab{}.
\newblock \showarticletitle{Fed-ComBat: A Generalized Federated Framework for Batch Effect Harmonization in Collaborative Studies}.
\newblock  (\bibinfo{date}{May} \bibinfo{year}{2023}).
\newblock
\urldef\tempurl%
\url{https://doi.org/10.1101/2023.05.24.542107}
\showDOI{\tempurl}


\bibitem[Singh et~al\mbox{.}(2022)]%
        {Singh2022}
\bibfield{author}{\bibinfo{person}{Nalini~M. Singh}, \bibinfo{person}{Jordan~B. Harrod}, \bibinfo{person}{Sandya Subramanian}, \bibinfo{person}{Mitchell Robinson}, \bibinfo{person}{Ken Chang}, \bibinfo{person}{Suheyla Cetin-Karayumak}, \bibinfo{person}{Adrian~Vasile Dalca}, \bibinfo{person}{Simon Eickhoff}, \bibinfo{person}{Michael Fox}, \bibinfo{person}{Loraine Franke}, \bibinfo{person}{Polina Golland}, \bibinfo{person}{Daniel Haehn}, \bibinfo{person}{Juan~Eugenio Iglesias}, \bibinfo{person}{Lauren~J. O’Donnell}, \bibinfo{person}{Yangming Ou}, \bibinfo{person}{Yogesh Rathi}, \bibinfo{person}{Shan~H. Siddiqi}, \bibinfo{person}{Haoqi Sun}, \bibinfo{person}{M.~Brandon Westover}, \bibinfo{person}{Susan Whitfield-Gabrieli}, {and} \bibinfo{person}{Randy~L. Gollub}.} \bibinfo{year}{2022}\natexlab{}.
\newblock \showarticletitle{How Machine Learning is Powering Neuroimaging to Improve Brain Health}.
\newblock \bibinfo{journal}{\emph{Neuroinformatics}} \bibinfo{volume}{20}, \bibinfo{number}{4} (\bibinfo{date}{March} \bibinfo{year}{2022}), \bibinfo{pages}{943–964}.
\newblock
\showISSN{1559-0089}
\urldef\tempurl%
\url{https://doi.org/10.1007/s12021-022-09572-9}
\showDOI{\tempurl}


\bibitem[Thompson et~al\mbox{.}(2014)]%
        {thompson2014enigma}
\bibfield{author}{\bibinfo{person}{Paul~M Thompson}, \bibinfo{person}{Jason~L Stein}, \bibinfo{person}{Sarah~E Medland}, \bibinfo{person}{Derrek~P Hibar}, \bibinfo{person}{Alejandro~Arias Vasquez}, \bibinfo{person}{Miguel~E Renteria}, \bibinfo{person}{Roberto Toro}, \bibinfo{person}{Neda Jahanshad}, \bibinfo{person}{Gunter Schumann}, \bibinfo{person}{Barbara Franke}, {et~al\mbox{.}}} \bibinfo{year}{2014}\natexlab{}.
\newblock \showarticletitle{The ENIGMA Consortium: large-scale collaborative analyses of neuroimaging and genetic data}.
\newblock \bibinfo{journal}{\emph{Brain imaging and behavior}}  \bibinfo{volume}{8} (\bibinfo{year}{2014}), \bibinfo{pages}{153--182}.
\newblock


\bibitem[Tibshirani(1996)]%
        {Tibshirani1996}
\bibfield{author}{\bibinfo{person}{Robert Tibshirani}.} \bibinfo{year}{1996}\natexlab{}.
\newblock \showarticletitle{Regression Shrinkage and Selection Via the Lasso}.
\newblock \bibinfo{journal}{\emph{Journal of the Royal Statistical Society: Series B (Methodological)}} \bibinfo{volume}{58}, \bibinfo{number}{1} (\bibinfo{date}{Jan.} \bibinfo{year}{1996}), \bibinfo{pages}{267–288}.
\newblock
\showISSN{2517-6161}
\urldef\tempurl%
\url{https://doi.org/10.1111/j.2517-6161.1996.tb02080.x}
\showDOI{\tempurl}


\bibitem[Wang et~al\mbox{.}(2023)]%
        {wang2023yolov7}
\bibfield{author}{\bibinfo{person}{Chien-Yao Wang}, \bibinfo{person}{Alexey Bochkovskiy}, {and} \bibinfo{person}{Hong-Yuan~Mark Liao}.} \bibinfo{year}{2023}\natexlab{}.
\newblock \showarticletitle{YOLOv7: Trainable bag-of-freebies sets new state-of-the-art for real-time object detectors}. In \bibinfo{booktitle}{\emph{Proceedings of the IEEE/CVF Conference on Computer Vision and Pattern Recognition}}. \bibinfo{pages}{7464--7475}.
\newblock


\bibitem[Wang et~al\mbox{.}(2018)]%
        {Wang2018}
\bibfield{author}{\bibinfo{person}{Qi Wang}, \bibinfo{person}{Lei Guo}, \bibinfo{person}{Paul~M. Thompson}, \bibinfo{person}{Clifford~R. Jack}, \bibinfo{person}{Hiroko Dodge}, \bibinfo{person}{Liang Zhan}, {and} \bibinfo{person}{Jiayu Zhou}.} \bibinfo{year}{2018}\natexlab{}.
\newblock \showarticletitle{The Added Value of Diffusion-Weighted MRI-Derived Structural Connectome in Evaluating Mild Cognitive Impairment: A Multi-Cohort Validation1}.
\newblock \bibinfo{journal}{\emph{Journal of Alzheimer’s Disease}} \bibinfo{volume}{64}, \bibinfo{number}{1} (\bibinfo{date}{June} \bibinfo{year}{2018}), \bibinfo{pages}{149–169}.
\newblock
\showISSN{1875-8908}
\urldef\tempurl%
\url{https://doi.org/10.3233/jad-171048}
\showDOI{\tempurl}


\bibitem[Wang et~al\mbox{.}(2017)]%
        {Wang2017}
\bibfield{author}{\bibinfo{person}{Qi Wang}, \bibinfo{person}{Mengying Sun}, \bibinfo{person}{Liang Zhan}, \bibinfo{person}{Paul Thompson}, \bibinfo{person}{Shuiwang Ji}, {and} \bibinfo{person}{Jiayu Zhou}.} \bibinfo{year}{2017}\natexlab{}.
\newblock \showarticletitle{Multi-Modality Disease Modeling via Collective Deep Matrix Factorization}. In \bibinfo{booktitle}{\emph{Proceedings of the 23rd ACM SIGKDD International Conference on Knowledge Discovery and Data Mining}} \emph{(\bibinfo{series}{KDD ’17})}. \bibinfo{publisher}{ACM}.
\newblock
\urldef\tempurl%
\url{https://doi.org/10.1145/3097983.3098164}
\showDOI{\tempurl}


\bibitem[Wang et~al\mbox{.}(2016)]%
        {Wang2016}
\bibfield{author}{\bibinfo{person}{Qi Wang}, \bibinfo{person}{Liang Zhan}, \bibinfo{person}{Paul~M. Thompson}, \bibinfo{person}{Hiroko~H. Dodge}, {and} \bibinfo{person}{Jiayu Zhou}.} \bibinfo{year}{2016}\natexlab{}.
\newblock \showarticletitle{Discriminative fusion of multiple brain networks for early mild cognitive impairment detection}. In \bibinfo{booktitle}{\emph{2016 IEEE 13th International Symposium on Biomedical Imaging (ISBI)}}. \bibinfo{pages}{568--572}.
\newblock
\urldef\tempurl%
\url{https://doi.org/10.1109/ISBI.2016.7493332}
\showDOI{\tempurl}


\bibitem[Welten et~al\mbox{.}(2022)]%
        {Welten2022}
\bibfield{author}{\bibinfo{person}{Sascha Welten}, \bibinfo{person}{Yongli Mou}, \bibinfo{person}{Laurenz Neumann}, \bibinfo{person}{Mehrshad Jaberansary}, \bibinfo{person}{Yeliz Yediel~Ucer}, \bibinfo{person}{Toralf Kirsten}, \bibinfo{person}{Stefan Decker}, {and} \bibinfo{person}{Oya Beyan}.} \bibinfo{year}{2022}\natexlab{}.
\newblock \showarticletitle{A Privacy-Preserving Distributed Analytics Platform for Health Care Data}.
\newblock \bibinfo{journal}{\emph{Methods of Information in Medicine}} \bibinfo{volume}{61}, \bibinfo{number}{S 01} (\bibinfo{date}{Jan.} \bibinfo{year}{2022}), \bibinfo{pages}{e1–e11}.
\newblock
\showISSN{2511-705X}
\urldef\tempurl%
\url{https://doi.org/10.1055/s-0041-1740564}
\showDOI{\tempurl}


\bibitem[Wirth et~al\mbox{.}(2021)]%
        {Wirth2021}
\bibfield{author}{\bibinfo{person}{Felix~Nikolaus Wirth}, \bibinfo{person}{Thierry Meurers}, \bibinfo{person}{Marco Johns}, {and} \bibinfo{person}{Fabian Prasser}.} \bibinfo{year}{2021}\natexlab{}.
\newblock \showarticletitle{Privacy-preserving data sharing infrastructures for medical research: systematization and comparison}.
\newblock \bibinfo{journal}{\emph{BMC Medical Informatics and Decision Making}} \bibinfo{volume}{21}, \bibinfo{number}{1} (\bibinfo{date}{Aug.} \bibinfo{year}{2021}).
\newblock
\showISSN{1472-6947}
\urldef\tempurl%
\url{https://doi.org/10.1186/s12911-021-01602-x}
\showDOI{\tempurl}


\bibitem[Xie et~al\mbox{.}(2017)]%
        {xie2017privacy}
\bibfield{author}{\bibinfo{person}{Liyang Xie}, \bibinfo{person}{Inci~M Baytas}, \bibinfo{person}{Kaixiang Lin}, {and} \bibinfo{person}{Jiayu Zhou}.} \bibinfo{year}{2017}\natexlab{}.
\newblock \showarticletitle{Privacy-preserving distributed multi-task learning with asynchronous updates}. In \bibinfo{booktitle}{\emph{Proceedings of the 23rd ACM SIGKDD international conference on knowledge discovery and data mining}}. \bibinfo{pages}{1195--1204}.
\newblock


\bibitem[Xie et~al\mbox{.}(2018)]%
        {xie2018differentially}
\bibfield{author}{\bibinfo{person}{Liyang Xie}, \bibinfo{person}{Kaixiang Lin}, \bibinfo{person}{Shu Wang}, \bibinfo{person}{Fei Wang}, {and} \bibinfo{person}{Jiayu Zhou}.} \bibinfo{year}{2018}\natexlab{}.
\newblock \showarticletitle{Differentially private generative adversarial network}.
\newblock \bibinfo{journal}{\emph{arXiv preprint arXiv:1802.06739}} (\bibinfo{year}{2018}).
\newblock


\bibitem[Yan et~al\mbox{.}(2023)]%
        {Yan2023}
\bibfield{author}{\bibinfo{person}{Zhiyu Yan}, \bibinfo{person}{Kori~S. Zachrison}, \bibinfo{person}{Lee~H. Schwamm}, \bibinfo{person}{Juan~J. Estrada}, {and} \bibinfo{person}{Rui Duan}.} \bibinfo{year}{2023}\natexlab{}.
\newblock \showarticletitle{A privacy-preserving and computation-efficient federated algorithm for generalized linear mixed models to analyze correlated electronic health records data}.
\newblock \bibinfo{journal}{\emph{PLOS ONE}} \bibinfo{volume}{18}, \bibinfo{number}{1} (\bibinfo{date}{Jan.} \bibinfo{year}{2023}), \bibinfo{pages}{e0280192}.
\newblock
\showISSN{1932-6203}
\urldef\tempurl%
\url{https://doi.org/10.1371/journal.pone.0280192}
\showDOI{\tempurl}


\bibitem[Yang et~al\mbox{.}(2021)]%
        {Yang}
\bibfield{author}{\bibinfo{person}{Jingkang Yang}, \bibinfo{person}{Kaiyang Zhou}, \bibinfo{person}{Yixuan Li}, {and} \bibinfo{person}{Ziwei Liu}.} \bibinfo{year}{2021}\natexlab{}.
\newblock \bibinfo{title}{Generalized Out-of-Distribution Detection: A Survey}.
\newblock
\newblock
\urldef\tempurl%
\url{https://doi.org/10.48550/ARXIV.2110.11334}
\showDOI{\tempurl}


\bibitem[Zahr et~al\mbox{.}(2009)]%
        {zahr2009problem}
\bibfield{author}{\bibinfo{person}{Natalie~M Zahr}, \bibinfo{person}{Torsten Rohlfing}, \bibinfo{person}{Adolf Pfefferbaum}, {and} \bibinfo{person}{Edith~V Sullivan}.} \bibinfo{year}{2009}\natexlab{}.
\newblock \showarticletitle{Problem solving, working memory, and motor correlates of association and commissural fiber bundles in normal aging: a quantitative fiber tracking study}.
\newblock \bibinfo{journal}{\emph{Neuroimage}} \bibinfo{volume}{44}, \bibinfo{number}{3} (\bibinfo{year}{2009}), \bibinfo{pages}{1050--1062}.
\newblock


\bibitem[Zeng et~al\mbox{.}(2023)]%
        {Fedlab}
\bibfield{author}{\bibinfo{person}{Dun Zeng}, \bibinfo{person}{Siqi Liang}, \bibinfo{person}{Xiangjing Hu}, \bibinfo{person}{Hui Wang}, {and} \bibinfo{person}{Zenglin Xu}.} \bibinfo{year}{2023}\natexlab{}.
\newblock \showarticletitle{FedLab: A Flexible Federated Learning Framework}.
\newblock \bibinfo{journal}{\emph{Journal of Machine Learning Research}} \bibinfo{volume}{24}, \bibinfo{number}{100} (\bibinfo{year}{2023}), \bibinfo{pages}{1--7}.
\newblock
\urldef\tempurl%
\url{http://jmlr.org/papers/v24/22-0440.html}
\showURL{%
\tempurl}


\bibitem[Zhan et~al\mbox{.}(2014)]%
        {zhan2014understanding}
\bibfield{author}{\bibinfo{person}{Liang Zhan}, \bibinfo{person}{Neda Jahanshad}, \bibinfo{person}{Yan Jin}, \bibinfo{person}{Talia~M Nir}, \bibinfo{person}{Cassandra~D Leonardo}, \bibinfo{person}{Matt~A Bernstein}, \bibinfo{person}{B Borowski}, \bibinfo{person}{Clifford~R Jack}, {and} \bibinfo{person}{Paul~M Thompson}.} \bibinfo{year}{2014}\natexlab{}.
\newblock \showarticletitle{Understanding scanner upgrade effects on brain integrity \& connectivity measures}. In \bibinfo{booktitle}{\emph{2014 IEEE 11th International Symposium on Biomedical Imaging (ISBI)}}. IEEE, \bibinfo{pages}{234--237}.
\newblock


\bibitem[Zhan et~al\mbox{.}(2015a)]%
        {Zhan2015}
\bibfield{author}{\bibinfo{person}{Liang Zhan}, \bibinfo{person}{Yashu Liu}, \bibinfo{person}{Yalin Wang}, \bibinfo{person}{Jiayu Zhou}, \bibinfo{person}{Neda Jahanshad}, \bibinfo{person}{Jieping Ye}, {and} \bibinfo{person}{Paul~M. Thompson}.} \bibinfo{year}{2015}\natexlab{a}.
\newblock \showarticletitle{Boosting brain connectome classification accuracy in Alzheimer’s disease using higher-order singular value decomposition}.
\newblock \bibinfo{journal}{\emph{Frontiers in Neuroscience}}  \bibinfo{volume}{9} (\bibinfo{date}{July} \bibinfo{year}{2015}).
\newblock
\showISSN{1662-453X}
\urldef\tempurl%
\url{https://doi.org/10.3389/fnins.2015.00257}
\showDOI{\tempurl}


\bibitem[Zhan et~al\mbox{.}(2015b)]%
        {Zhan2015-2}
\bibfield{author}{\bibinfo{person}{Liang Zhan}, \bibinfo{person}{Jiayu Zhou}, \bibinfo{person}{Yalin Wang}, \bibinfo{person}{Yan Jin}, \bibinfo{person}{Neda Jahanshad}, \bibinfo{person}{Gautam Prasad}, \bibinfo{person}{Talia~M. Nir}, \bibinfo{person}{Cassandra~D. Leonardo}, \bibinfo{person}{Jieping Ye}, \bibinfo{person}{Paul~M. Thompson}, {and} \bibinfo{person}{for~the Alzheimer’s Disease Neuroimaging~Initiative}.} \bibinfo{year}{2015}\natexlab{b}.
\newblock \showarticletitle{Comparison of nine tractography algorithms for detecting abnormal structural brain networks in Alzheimer’s disease}.
\newblock \bibinfo{journal}{\emph{Frontiers in Aging Neuroscience}}  \bibinfo{volume}{7} (\bibinfo{date}{April} \bibinfo{year}{2015}).
\newblock
\showISSN{1663-4365}
\urldef\tempurl%
\url{https://doi.org/10.3389/fnagi.2015.00048}
\showDOI{\tempurl}


\bibitem[Zhou et~al\mbox{.}(2012)]%
        {zhou2012modeling}
\bibfield{author}{\bibinfo{person}{Jiayu Zhou}, \bibinfo{person}{Jun Liu}, \bibinfo{person}{Vaibhav~A Narayan}, {and} \bibinfo{person}{Jieping Ye}.} \bibinfo{year}{2012}\natexlab{}.
\newblock \showarticletitle{Modeling disease progression via fused sparse group lasso}. In \bibinfo{booktitle}{\emph{Proceedings of the 18th ACM SIGKDD international conference on Knowledge discovery and data mining}}. \bibinfo{pages}{1095--1103}.
\newblock


\bibitem[Zhou et~al\mbox{.}(2013)]%
        {zhou2013modeling}
\bibfield{author}{\bibinfo{person}{Jiayu Zhou}, \bibinfo{person}{Jun Liu}, \bibinfo{person}{Vaibhav~A Narayan}, \bibinfo{person}{Jieping Ye}, \bibinfo{person}{Alzheimer's Disease~Neuroimaging Initiative}, {et~al\mbox{.}}} \bibinfo{year}{2013}\natexlab{}.
\newblock \showarticletitle{Modeling disease progression via multi-task learning}.
\newblock \bibinfo{journal}{\emph{NeuroImage}}  \bibinfo{volume}{78} (\bibinfo{year}{2013}), \bibinfo{pages}{233--248}.
\newblock


\bibitem[Zhou et~al\mbox{.}(2011)]%
        {zhou2011multi}
\bibfield{author}{\bibinfo{person}{Jiayu Zhou}, \bibinfo{person}{Lei Yuan}, \bibinfo{person}{Jun Liu}, {and} \bibinfo{person}{Jieping Ye}.} \bibinfo{year}{2011}\natexlab{}.
\newblock \showarticletitle{A multi-task learning formulation for predicting disease progression}. In \bibinfo{booktitle}{\emph{Proceedings of the 17th ACM SIGKDD international conference on Knowledge discovery and data mining}}. \bibinfo{pages}{814--822}.
\newblock


\end{thebibliography}

\appendix

\end{document}